\documentclass[11pt]{article}

\usepackage[preprint]{acl}
\usepackage{pifont}
\usepackage{xcolor}
\usepackage{booktabs}

\usepackage{times}
\usepackage{latexsym}
\usepackage[T1]{fontenc}
\usepackage[utf8]{inputenc}
\usepackage{microtype}
\usepackage{inconsolata}
\usepackage{graphicx}
\graphicspath{{figures/}}
\usepackage{booktabs}
\usepackage{multirow}
\usepackage{array}
\usepackage{amsmath}
\usepackage{amssymb}
\usepackage{xcolor}
\usepackage{subcaption}
\usepackage{pifont}
\usepackage{placeins}

\title{Reading What Images Imply: Visual Opinion Mining\\ in Vision-Language Models}

\author{
  \textbf{Huiyao Chen\textsuperscript{1,2}}\thanks{Equal contribution.},
  \textbf{Bonian Jia\textsuperscript{1,2}}\footnotemark[1],
  \textbf{Yi Yang\textsuperscript{1,2}},
  \textbf{Yumeng Fu\textsuperscript{3}},
  \\
  \textbf{Meishan Zhang\textsuperscript{1}}\thanks{Corresponding author: Meishan Zhang.},
  \textbf{Baotian Hu\textsuperscript{1,2}},
  \textbf{Min Zhang\textsuperscript{1,2}}
  \\
  \textsuperscript{1} Institute of Computing and Intelligence, Harbin Institute of Technology (Shenzhen)
  \\
  \textsuperscript{2} Shenzhen Loop Area Institute (SLAI)
  \quad
  \textsuperscript{3} Harbin Institute of Technology
  \\
  \texttt{\{chenhy1018,mason.zms\}@gmail.com}, \quad
  \texttt{bonianjia@slai.edu.cn}
}

\hypersetup{
  pdftitle={Reading What Images Imply: Visual Opinion Mining in Vision-Language Models},
  pdfauthor={Huiyao Chen, Bonian Jia, Yi Yang, Yumeng Fu, Meishan Zhang, Baotian Hu, Min Zhang}
}

\begin{document}
\begin{NoHyper}
\maketitle
\end{NoHyper}
\begin{abstract}
People in the same image can express different attitudes toward different targets.
Capturing these distinctions helps image descriptions convey how participants respond to a scene.
We introduce Visual Opinion Mining, extending structured opinion mining to images without accompanying text.
Each opinion links a holder, target, expression, and polarity to supporting visual evidence.
We annotate 5,239 images with 7,406 opinions and design four diagnostic probes.
By varying spatial and textual guidance, the probes separate interpretation, localization, and discovery.
Our analysis of vision-language models shows that stronger opinion descriptions can accompany weaker evidence grounding.
Expressive cues remain difficult to locate even when their descriptions are supplied.
Adaptation improves discovery more consistently than cue localization, while strong single-holder performance can mask missed attitudes in multi-holder scenes.
Refining evidence boxes improves grounded discovery while keeping opinion text fixed.
These findings identify expressive-cue grounding and multi-holder coverage as distinct challenges in recovering structured opinions from images.
\end{abstract}

\section{Introduction}
\label{sec:intro}

\begin{figure}[t]
  \centering
  \includegraphics[width=\columnwidth,pagebox=cropbox]{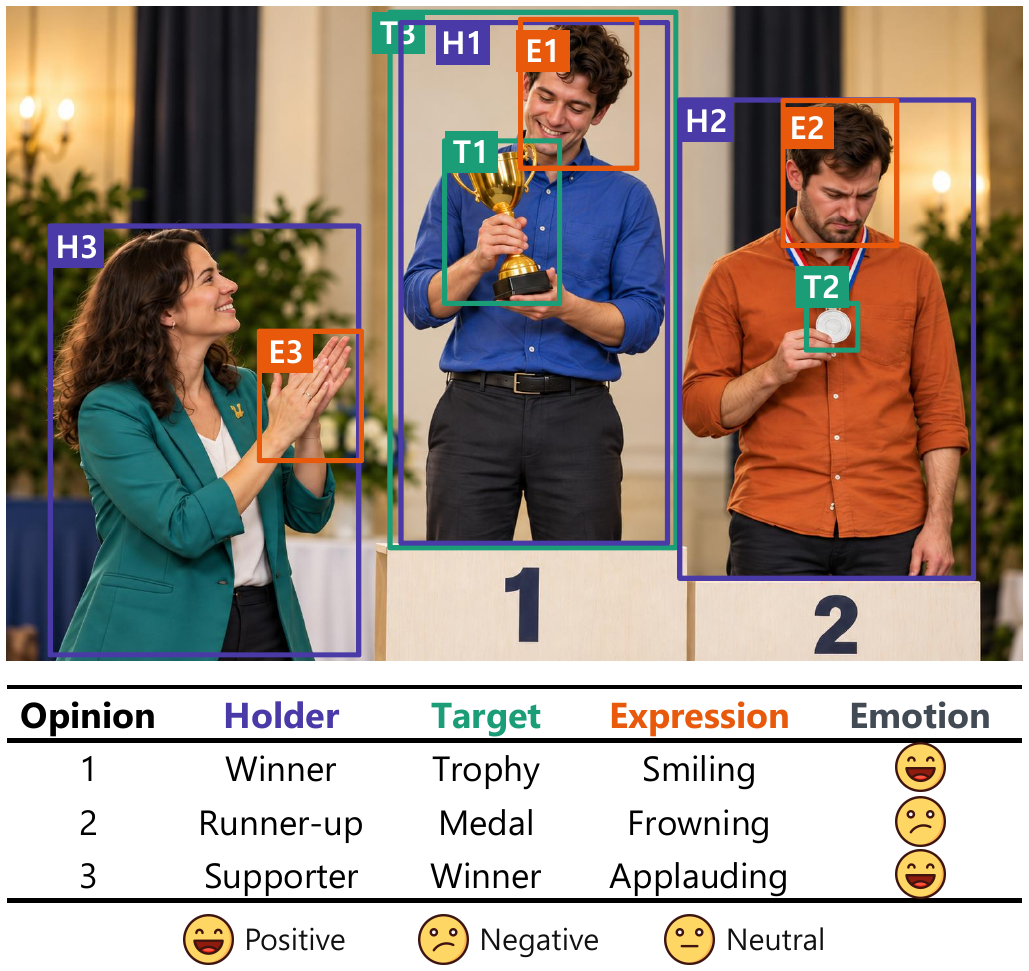}
  \caption{Multiple visual opinions in one scene.
  Purple, green, and orange boxes mark the Holder, Target, and Expression of each numbered opinion.
  Icons indicate Emotion polarity.
  The winner is the Holder of opinion 1 and the Target of opinion 3.}
  \label{fig:task_example}
\end{figure}

People can read attitudes from a photograph without a word of dialogue.
Consider the award ceremony in Figure~\ref{fig:task_example}.
The winner smiles at the trophy, the runner-up frowns at the medal, and a supporter applauds the winner.
These cues suggest satisfaction, disappointment, and approval toward different targets.
Describing the scene as an award ceremony names the event but omits how its participants respond.
Calling it a happy occasion also overlooks the apparent disappointment.
Preserving these differences matters when conveying what a photograph shows about the people in it.
Human emotion judgments draw on body cues as well as faces~\citep{aviezer2012body}.
Visual-assistance research likewise identifies contextual understanding as a concern for users who rely on generated descriptions~\citep{karamolegkou2025assistants}.

Opinion mining is an established NLP task that identifies attitudes and their targets.
Structured sentiment analysis links opinion holders, targets, expressions, and polarities~\citep{barnes2021structured,barnes2022semeval}.
In the award scene, the winner holds an attitude toward the trophy and receives approval from the supporter.
We extend this structure to images through \textbf{Visual Opinion Mining}.
Given only an image, a model must describe these roles and locate their visual evidence.

This kind of interpretation is a concrete problem in multimodal reasoning~\citep{zellers2019vcr,park2020visualcomet}.
Vision-language models support image description and question answering through natural language~\citep{li2023blip2,liu2023visualinstruction,bai2025qwen25vltechnicalreport}.
Social reasoning benchmarks extend evaluation to interactions and mental-state inference~\citep{zadeh2019socialiq,jin2024mmtom,niu2026read}.
Other work tests whether gaze and pose reveal hidden scene information~\citep{balamurugan2025spot}.
For participant attitudes, however, describing a visible cue is only part of the problem.
A smile alone does not identify what the person is pleased about.
A plausible interpretation also does not specify where its supporting evidence lies.
We therefore ask: \emph{Can models discover participant attitudes and their targets, and locate the visual evidence supporting each interpretation?}

Existing tasks provide useful components.
Visual sentiment and emotion recognition assign affective labels to images or people~\citep{you2015robust,kosti2019context,lee2019caer,mittal2020emoticon}.
Phrase grounding locates a supplied description~\citep{yu2018mattnet,kamath2021mdetr}, while grounded generation connects language to regions or masks~\citep{you2024ferret,rasheed2024glamm,lai2024lisa}.
Aspect-based multimodal opinion mining uses textual aspects to guide visual analysis~\citep{ling2022vision,zhou2023aom}, including object grounding in CORSA~\citep{liu2025corsa}.
Our setting supplies only the image.
The model must decide which attitudes to describe and identify the supporting regions.

Evidence-based visual inference offers a related perspective.
Sherlock evaluates localization of evidence for abductive inferences~\citep{hessel2022sherlock}.
Seeing Culture pairs cultural question answering with segmentation of relevant artifacts~\citep{satar2025seeing}.
We distinguish the person, the target, and the expressive cue.
Locating the winner alone does not identify the smile that supports satisfaction.
This distinction lets us assess attitude descriptions together with their supporting visual regions.

Our task predicts a set of grounded opinion quadruples from an image alone.
Each opinion specifies a Holder, Target, Expression, and Emotion polarity.
The Holder identifies who expresses the attitude, and the Target specifies what it concerns.
The Expression describes the visible behavior that supports the interpretation.
The first three roles receive bounding boxes.
A set of opinions preserves different reactions within one scene, including several attitudes toward the same target.
The annotations record visually supported interpretations rather than verified mental states.

The task combines interpretation, localization, and discovery.
Attitudes may be implicit in posture or gaze.
A scene may contain several holders, and the number of opinions is unknown.
We define four probes to separate these demands.
Probe A asks for a role description given its region.
Probe B asks for a region given its phrase.
Probe C supplies an opinion sentence and requires extraction and grounding.
Probe D requires discovery from the image alone.
These probes test which errors persist under textual or spatial guidance.

We manually annotate 5,239 images with 7,406 opinions, reserving 100 images for adaptation.
We evaluate seven open vision-language models and three single-task models, plus two proprietary models on a paired test subset.
The strongest zero-shot open model reaches 61.24 semantic discovery F1, but only 15.50 joint F1 when emotion and grounding are required.
Even with the Expression phrase supplied, zero-shot localization accuracy reaches only 15.69\%.
Adaptation improves semantic discovery for all six adapted models, while Expression localization remains low.

Our contributions are:
\begin{itemize}
  \item An image-only formulation of Visual Opinion Mining that links participant attitudes to their targets and visual evidence.
  \item A manually annotated corpus with four probes for interpreting regions, locating evidence, and discovering opinions.
  \item An empirical analysis identifying expressive-cue grounding and multi-holder coverage as persistent difficulties in recovering structured opinions from visual evidence.
\end{itemize}

Our code and dataset are publicly available at \href{https://github.com/jbnhandsome/visual-opinion-mining}{github/jbnhandsome/visual-opinion-mining} to facilitate future research.

\section{Related Work}
\label{sec:related}

\subsection{Opinion Mining across Modalities}

Textual opinion mining represents sentiment through aspects, expressions, and their relations~\citep{pontiki2016semeval,xu2020position,peng2020knowing,xu2026rationale}.
Aspect sentiment quadruples include aspect categories~\citep{cai2021aspect,zhang2021aspect}.
Structured sentiment analysis links holders, targets, expressions, and polarities~\citep{barnes2021structured,barnes2022semeval}.
Multimodal sentiment research combines images with posts~\citep{niu2016sentiment} or speech with facial expressions and gestures~\citep{zadeh2018multimodal,hazarika2020misa}.
Opinion expression identification also integrates text and speech~\citep{jia2024llmdriven}.
Aspect-based approaches select visual information relevant to textual aspects~\citep{yu2019adapting,ling2022vision,zhou2023aom}.
CORSA also grounds related visual objects with bounding boxes~\citep{liu2025corsa}.
These approaches use text, speech, or supplied aspects to guide prediction.
We extend structured opinion mining to image-only discovery, generating role descriptions and grounding each role in visual evidence.

\subsection{Visual Social and Affective Understanding}

Visual reasoning benchmarks examine information beyond object identity.
VCR evaluates answers and rationales, while VisualCOMET predicts surrounding events and human intentions~\citep{zellers2019vcr,park2020visualcomet}.
Social-IQ, MMToM-QA, and Read the Room examine social situations and mental-state inference~\citep{zadeh2019socialiq,jin2024mmtom,niu2026read}.
Spot The Ball tests whether behavioral cues reveal a hidden object location~\citep{balamurugan2025spot}.
Visual sentiment methods predict affective labels using image features and region attention~\citep{you2015robust,campos2017pixels,you2017visual,zhao2021affective}.
Body context influences perceived facial affect~\citep{aviezer2012body}, and emotion models combine person appearance with contextual cues~\citep{kosti2019context,lee2019caer,mittal2020emoticon}.
We discover multiple participant attitudes in a single image and ground the person, target, and expressive cue for each one.

\subsection{Grounding Visual Evidence}

Phrase-conditioned grounding locates regions described by text~\citep{yu2018mattnet,yang2020improving,kamath2021mdetr,liu2024groundingdino}.
Grounded generation associates language with image regions~\citep{peng2023kosmos,chen2023shikra,you2024ferret,wu2024grit} or segmentation masks~\citep{rasheed2024glamm}.
LISA also segments objects from implicit queries that require reasoning~\citep{lai2024lisa}.
Beyond grounding, evaluation examines natural-language explanations~\citep{kayser2021evil} and object hallucinations~\citep{rohrbach2018hallucination,li2023pope}.
Sherlock explicitly evaluates evidence localization for abductive inferences~\citep{hessel2022sherlock}.
Seeing Culture combines cultural question answering with segmentation of relevant artifacts~\citep{satar2025seeing}.
Our four probes test how opinion descriptions align with visual evidence under different forms of textual and spatial guidance.

\section{Task and Diagnostic Evaluation}
\label{sec:task}

\subsection{Visual Opinion Mining}
\label{sec:formulation}

Given only an image $I$, a model predicts a set of opinions $\mathcal{O}(I)=\{o_1,\ldots,o_n\}$.
The number of opinions $n$ is unknown.
Following structured sentiment analysis~\citep{barnes2021structured,barnes2022semeval}, we represent each opinion as
\begin{equation}
o_i=\big((h_i,\mathbf{b}_i^h),(t_i,\mathbf{b}_i^t),(e_i,\mathbf{b}_i^e),s_i\big).
\end{equation}
The phrases $h_i$, $t_i$, and $e_i$ describe the Holder, Target, and Expression.
The Holder is a visible person or group, the Target is what the attitude concerns, and the Expression describes the supporting behavior.
The label $s_i$ is Positive, Negative, or Neutral.
Each grounded role has one box $\mathbf{b}_i^x$.
Expression boxes may cover a face, a gesture, or a whole person when posture carries the cue.

\subsection{Four Diagnostic Probes}
\label{sec:subtasks}

\begin{figure*}[t]
  \centering
  \includegraphics[width=\textwidth]{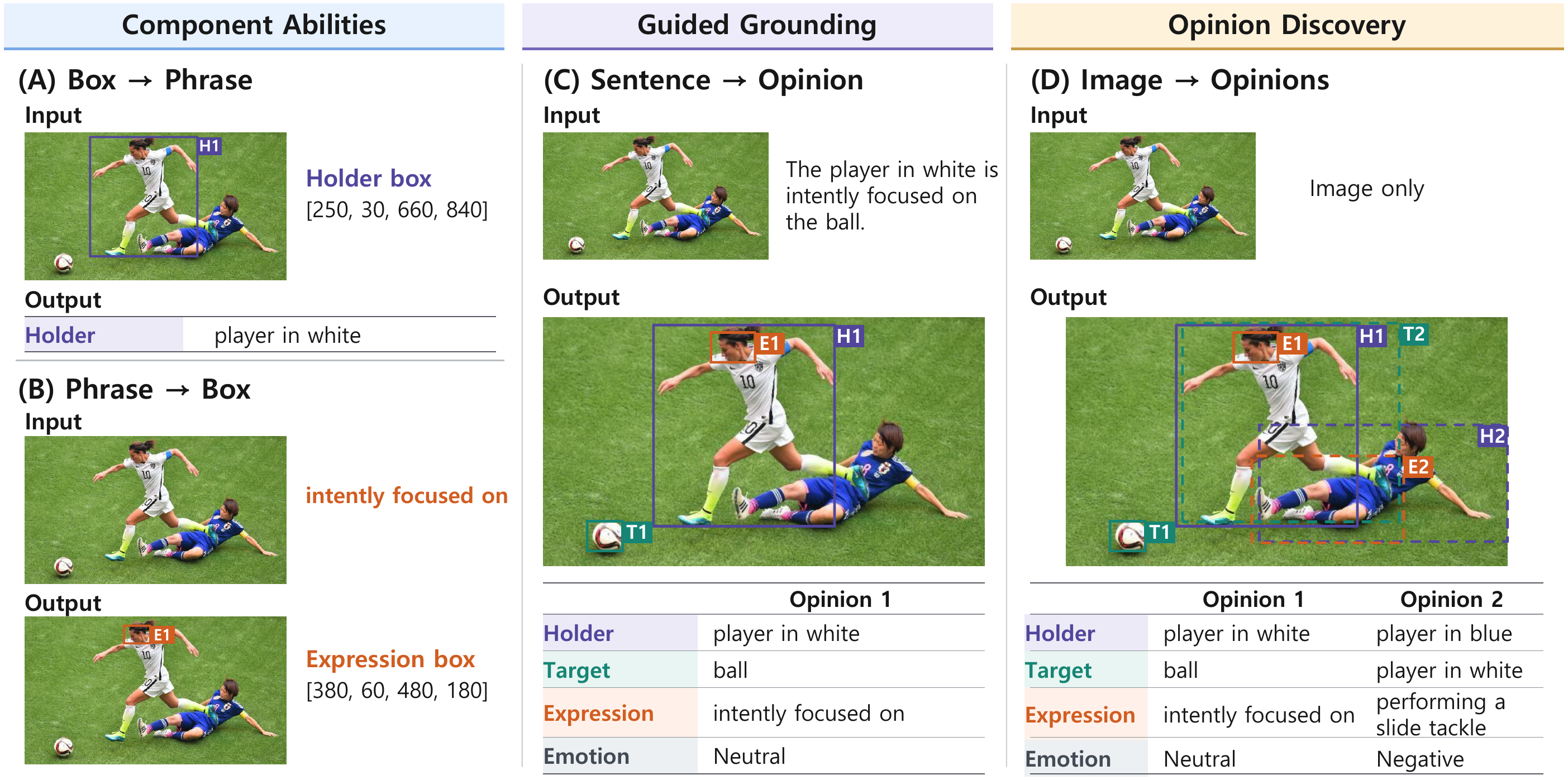}
  \caption{Four probes under different guidance conditions.
  A and B test component abilities, C tests guided grounding, and D tests opinion discovery.
  All receive the image and are evaluated independently.
  Purple, green, and orange mark Holder, Target, and Expression.
  Dashed boxes in D distinguish the second opinion.
  Box coordinates are normalized to 0--1000.}
  \label{fig:subtasks}
\end{figure*}

Figure~\ref{fig:subtasks} groups the four probes by the abilities they test.
Probes A and B test component abilities: interpreting a supplied region and locating a supplied phrase.
Probe C combines extraction and localization under sentence guidance.
Probe D adds the need to discover opinions from the image alone.

\paragraph{A: Region interpretation.}
Given the image, a reference box, and its opinion role, predict the corresponding phrase.
This tests interpretation when the relevant region is supplied.

\paragraph{B: Evidence localization.}
Given the image and a role phrase, predict its box.
This tests localization when the description is supplied.

\paragraph{C: Guided opinion grounding.}
Given the image and an opinion sentence, extract the quadruple and predict its role boxes.
This tests extraction and localization for a specified opinion.

\paragraph{D: Image-only opinion discovery.}
Given only the image, predict all supported opinion quadruples and their boxes.
The model must decide how many opinions to report.

The probes are evaluated independently.
Supplying a region, phrase, or sentence separates interpretation and localization errors from failures to discover an opinion.
Each probe measures performance under a specified form of guidance.

\subsection{Evaluation Metrics}
\label{sec:metrics}

\paragraph{Component and guided scores.}
Probe A reports mean BGE embedding cosine similarity~\citep{xiao2023cpack} between predicted and reference phrases.
Probe B reports box accuracy at IoU $\geq0.5$ by role and over eligible queries.
Probe C requires all three phrases to reach cosine $0.5$, exact Emotion, and mean IoU $\geq0.5$ over available reference boxes.
For available roles $R(o)$, the spatial gate is
\begin{equation}
\frac{1}{|R(o)|}\sum_{r\in R(o)}\operatorname{IoU}(\hat{\mathbf b}^{r},\mathbf b^{r})\geq0.5.
\end{equation}
We report the joint score and its individual gates.
Spatial scores exclude missing reference boxes.
Omitted predictions are counted as errors.

\paragraph{Discovery scores.}
Probe D matches predicted and reference opinions one-to-one within each image.
Semantic F1 matches opinions when all three role phrases reach cosine $0.5$.
Joint opinion F1 adds exact Emotion and the mean-IoU grounding criterion.
The spatial gate averages overlap across available role boxes.
Role-level scores measure localization for each type of evidence.
Conditional box accuracy uses semantically matched elements with reference boxes.
Grounded recall divides correctly localized semantic matches by all available reference boxes, including undiscovered elements.

Invalid outputs count as incorrect, and parse rate is an output diagnostic.
Scores are multiplied by 100 throughout.
We compare models within each metric and guidance condition.
Paired differences use 95\% image-bootstrap intervals with 2,000 resamples.
Reference precision measures agreement with the annotated opinions.
Appendix~\ref{sec:appendix_metrics} gives the full metric definitions.

\section{Dataset Construction}
\label{sec:dataset}

\paragraph{Source and annotation.}
\label{sec:source}
Six annotators manually annotated 5,239 images from an EMOTIC-derived pool of human scenes~\citep{kosti2019context}.
Four annotators labeled the first 3,000 images, and the other two reviewed them.
These two reviewers labeled the remaining 2,239 images.
Reviewers accepted 85\% of the first batch on first review and about 80\% of 200 randomly sampled images from the second batch.
Rejected annotations were revised.
Each opinion contains three short role phrases, Emotion polarity, and one box per available grounded role.
Annotators write one sentence for each supported Holder--Target pair and mark its three role phrases.
One person can hold opinions toward several targets, and several people can react to the same target.
Each opinion links an attitude to a specific participant and target.
The Holder and Target boxes cover the visible entities.
The Expression box covers the visible cue, including the whole person for posture or body language.
Appendix~\ref{sec:appendix_guidelines} gives the full guidelines.

\paragraph{Adaptation and test split.}
\label{sec:split}
We reserve 100 images with 256 opinions for adaptation and use the remaining 5,139 images for testing.
The adaptation images are sampled with a fixed seed from multi-opinion images, stratified as 67 with two opinions and 33 with three or more.
This deliberately small split tests limited task adaptation.
Image identifiers are disjoint, and a dHash audit at distance $\leq6$ finds no near-duplicate pair across partitions.
Table~\ref{tab:dataset_summary} summarizes the split.

\begin{table}[t]
\centering
\small
\begin{tabular}{lrr}
\toprule
 & Adaptation & Test \\
\midrule
Images & 100 & 5,139 \\
Opinions & 256 & 7,150 \\
Target boxes & 244 & 6,796 \\
\bottomrule
\end{tabular}
\caption{Dataset split.
Every test opinion has Holder and Expression boxes.}
\label{tab:dataset_summary}
\end{table}

\paragraph{Reference coverage.}
\label{sec:bbox-audit}
All test opinions have Holder and Expression boxes.
The 354 opinions without Target boxes retain their text but are excluded from Target-box evaluation.
Probes A and B each contain 21,096 eligible role queries.
Median box area is 17.7\% of the image for Holder, 9.0\% for Target, and 2.0\% for Expression.
Appendix~\ref{sec:region_extent} describes cue regions and the effect of box size on IoU.

\paragraph{Opinion diversity.}
\label{sec:statistics}
The test set averages 1.39 opinions per image and contains up to 12.
It includes 1,327 multi-opinion images, of which 643 contain at least two distinct holders.
The grouping uses holder identity regardless of attitude polarity.
Emotion labels are 64.3\% Positive, 22.5\% Negative, and 13.1\% Neutral.
For exploratory analysis, GPT-5.5 labels 599 randomly sampled opinions from 375 test images.
It receives the image, annotated boxes, and an Expression-region close-up.
An opinion is implicit when its attitude relies on posture, gaze, action, or context without a readable facial expression.
In this sample, 76.0\% are labeled implicit.
A manual check of 200 labels agrees on 79.0\% of implicitness decisions.
We use these labels for descriptive comparisons, with detailed statistics in Appendix~\ref{sec:appendix_statistics}.

\section{Experiments}
\label{sec:experiments}

We organize the evaluation around three questions:
\textbf{RQ1:} How well can models discover and ground opinions from images alone?
\textbf{RQ2:} Which difficulties persist when regions, phrases, or opinion sentences are supplied?
\textbf{RQ3:} What does adaptation on 100 images improve?

\subsection{Experimental Setup}
\label{sec:setup}

We evaluate seven open VLMs from five families, listed in Table~\ref{tab:main_results}.
Grounding DINO-B~\citep{liu2024groundingdino} and OWLv2-B provide single-task references for Probe B.
Florence-2-L~\citep{xiao2024florence2} performs Probes A and B.
Open systems use all 5,139 test images on applicable probes.
This yields 21,096 queries each for A and B, 7,150 for C, and 5,139 for D.
Appendix~\ref{sec:api} reports two proprietary models on a paired subset of 100 test images.

We adapt six open VLMs, excluding Ovis2.5-2B, on 1,868 supervision instances from the 100 adaptation images.
All use language-side LoRA with rank 16, three epochs, learning rate $10^{-4}$, and effective batch size 16.
We evaluate the final checkpoint.
The adapted prompt also permits whole-person boxes when the cue is too small to isolate.
Models return structured JSON with boxes converted to a common coordinate representation.
Decoding is greedy except for zero-shot GLM, which samples with thinking enabled.
Adapted GLM answers directly, so its comparison includes a decoding change.
Appendix~\ref{sec:appendix_reproducibility} gives implementation details, and Appendix~\ref{sec:appendix_prompts} lists the prompts.

\begin{table}[t]
\centering
\footnotesize
\setlength{\tabcolsep}{2pt}
\begin{tabular*}{\columnwidth}{@{\extracolsep{\fill}}lrrrrr@{}}
\toprule
& \textbf{A} & \textbf{B} & \textbf{C} & \multicolumn{2}{c}{\textbf{D}} \\
\cmidrule(lr){5-6}
\textbf{Model} & BGE & Acc & Joint & \shortstack{Quad.\\F1} & \shortstack{Joint\\F1} \\
\midrule
\multicolumn{6}{l}{\textit{Zero-shot vision-language models}} \\
Qwen3.5-9B & 65.38 & 47.64 & 31.50 & 45.33 & 11.21 \\
Qwen3.8-27B & 64.27 & \textbf{48.03} & \textbf{38.29} & \textbf{61.24} & \textbf{15.50} \\
InternVL3.5-8B & 65.51 & 9.84 & 3.94 & 47.55 & 1.05 \\
Ovis2.5-9B & 66.94 & 33.67 & 20.99 & 56.06 & 7.42 \\
Ovis2.5-2B & 65.26 & 35.48 & 15.51 & 58.18 & 5.77 \\
GLM-4.1V-9B & 68.37 & 46.91 & 33.90 & 19.93 & 7.16 \\
MiniCPM-V-4.5 & \textbf{68.53} & 40.36 & 5.87 & 53.35 & 2.46 \\
\midrule
\multicolumn{6}{l}{\textit{Adapted on the 100 adaptation images (LoRA SFT)}} \\
Qwen3.5-9B~+SFT & 79.57 & 45.92 & 39.61 & 63.34 & 15.66 \\
Qwen3.8-27B~+SFT & \textbf{79.99} & \textbf{46.54} & \textbf{39.64} & 64.38 & 15.97 \\
InternVL3.5-8B~+SFT & 77.20 & 39.04 & 34.80 & 60.23 & 13.69 \\
Ovis2.5-9B~+SFT & 78.09 & 42.57 & 34.01 & \textbf{71.55} & \textbf{16.74} \\
GLM-4.1V-9B~+SFT & 79.16 & 46.46 & 39.05 & 68.95 & 16.54 \\
MiniCPM-V-4.5~+SFT & 79.47 & 42.70 & 28.69 & 70.26 & 12.14 \\
\bottomrule
\end{tabular*}
\caption{Full-test results.
A: mean BGE cosine.
B: box accuracy at IoU $\geq0.5$.
C: joint score.
D: semantic quadruple and joint opinion F1.
Scores are on a 0--100 scale.
Bold marks the best result within each block.
GLM denotes GLM-4.1V-9B-Thinking.}
\label{tab:main_results}
\end{table}

\subsection{RQ1: Image-only Opinion Discovery}
\label{sec:results}

\paragraph{Grounding changes which models perform best.}
Table~\ref{tab:main_results} reports all four probes, with Probe D addressing the complete task.
Qwen3.8-27B leads zero-shot semantic discovery with 61.24 F1, but reaches only 15.50 joint F1.
The strongest adapted model, Ovis2.5-9B, reaches 71.55 and 16.74 respectively.
Joint scoring adds exact Emotion and spatial grounding to semantic agreement.
These requirements also reverse some model rankings.
Ovis2.5-2B exceeds Qwen3.5-9B in semantic F1, at 58.18 versus 45.33.
Yet its joint F1 is about half as high, at 5.77 versus 11.21.
The ranking reversal shows why opinion descriptions and visual evidence need joint evaluation.

Appendix~\ref{sec:appendix_full_results} shows a persistent weakness in expressive-cue localization.
Zero-shot grounded Expression recall reaches at most 8.71\%.
Even among semantically matched elements, Expression box accuracy is only 2.80--11.77\% across the seven open VLMs.
Many semantically matched opinions thus have poorly localized expressive evidence.

\paragraph{High conditional scores can hide poor coverage.}
Zero-shot systems produce 1,481--15,689 opinions for 7,150 references.
GLM returns only 1,481 opinions and parseable output for 74.37\% of images.
It has the highest Emotion accuracy on matched opinions, at 87.91\%, yet the lowest semantic F1, at 19.93.
Its strong conditional score therefore coexists with substantial under-generation.
Qwen3.5-9B produces more than twice as many opinions as the reference set, but reaches only 45.33 semantic F1.
Opinion discovery requires both recovering annotated attitudes and selecting which interpretations to report.

\paragraph{Proprietary models show the same cue-level weakness.}
On the same 100 test images, GPT-5.5 and Gemini-3.6-Flash obtain semantic F1 of 57.08 and 61.58.
Their joint F1 falls to 6.03 and 9.07, with grounded Expression recall of 3.75\% and 3.12\%.
Qwen3.8-27B reaches 62.07 semantic F1 and 10.92 joint F1 on these images.
The two proprietary models also pass the Probe C phrase criterion on at least 99.38\% of opinions.
Their corresponding joint scores are only 20.00 and 28.75.
This gap persists under sentence guidance.

\subsection{RQ2: Errors under Guidance}
\label{sec:analysis}

\begin{figure}[t]
\centering
\includegraphics[width=\columnwidth]{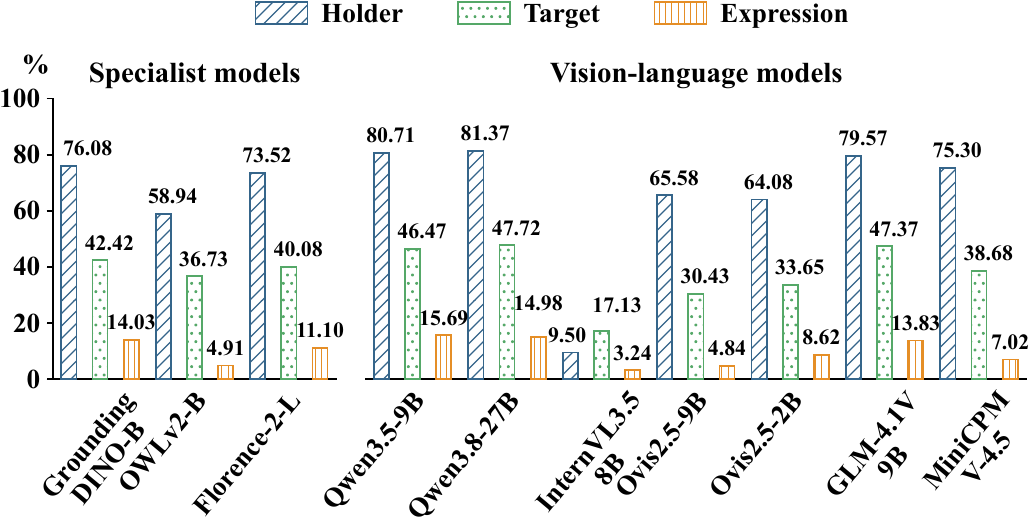}
\caption{Zero-shot phrase-to-box accuracy by role in Probe B, at IoU $\geq0.5$.
The two panels compare specialist grounders and VLMs on the full test set.
Expression is the least accurately localized role for every system.
Table~\ref{tab:probe_b} includes adapted results.}
\label{fig:role_summary}
\end{figure}

\paragraph{Knowing the phrase leaves the expressive cue hard to locate.}
Probe B supplies the role description.
Expression is still the least accurately localized role for all ten zero-shot systems in Figure~\ref{fig:role_summary}.
Qwen3.5-9B locates 80.71\% of Holders but only 15.69\% of Expressions.
Grounding DINO-B shows a similar contrast, at 76.08\% and 14.03\%.
Both VLMs and specialist grounders locate people much more reliably than their expressive cues.
Expression boxes also occupy a median of 2.0\% of the image, compared with 17.7\% for Holders.
Grounding an attitude requires locating these small or diffuse cues from their descriptions.

\paragraph{A sentence makes extraction easier than grounding.}
Probe C supplies the full opinion sentence.
Table~\ref{tab:guided_summary} separates its semantic, spatial, and emotion criteria.
For Qwen3.8-27B, all three phrases pass the semantic threshold on 99.15\% of opinions.
The mean-IoU gate passes on 48.01\%, Emotion on 79.06\%, and all criteria together on 38.29\%.
InternVL extracts the phrases on 96.48\% of opinions, but passes the spatial gate on only 4.92\%.
In contrast, MiniCPM has a phrase pass rate of 54.36\% and thus also struggles with extraction.
Role extraction and evidence grounding pose distinct challenges, illustrated in Appendix~\ref{sec:appendix_cases}.

\begin{table}[t]
\centering
\small
\setlength{\tabcolsep}{0pt}
\newlength{\guidedmodelwidth}
\settowidth{\guidedmodelwidth}{MiniCPM-V-4.5}
\newlength{\guidedmetricwidth}
\setlength{\guidedmetricwidth}{\dimexpr(\columnwidth-\guidedmodelwidth)/4\relax}
\newcommand{\guidedhead}[1]{\makebox[\guidedmetricwidth][c]{#1}}
\newcommand{\guidedvalue}[1]{\makebox[\guidedmetricwidth][c]{\makebox[2.25em][r]{#1}}}
\begin{tabular*}{\columnwidth}{@{}lrrrr@{}}
\toprule
Model & \guidedhead{All phrases} & \guidedhead{IoU gate} & \guidedhead{Emotion} & \guidedhead{Joint score} \\
\midrule
Qwen3.5-9B & \guidedvalue{94.50} & \guidedvalue{42.38} & \guidedvalue{78.08} & \guidedvalue{31.50} \\
Qwen3.8-27B & \guidedvalue{\textbf{99.15}} & \guidedvalue{\textbf{48.01}} & \guidedvalue{79.06} & \guidedvalue{\textbf{38.29}} \\
InternVL3.5-8B & \guidedvalue{96.48} & \guidedvalue{4.92} & \guidedvalue{83.69} & \guidedvalue{3.94} \\
Ovis2.5-9B & \guidedvalue{97.80} & \guidedvalue{26.10} & \guidedvalue{80.64} & \guidedvalue{20.99} \\
Ovis2.5-2B & \guidedvalue{98.22} & \guidedvalue{18.18} & \guidedvalue{84.83} & \guidedvalue{15.51} \\
GLM-4.1V-9B & \guidedvalue{98.28} & \guidedvalue{39.20} & \guidedvalue{\textbf{86.99}} & \guidedvalue{33.90} \\
MiniCPM-V-4.5 & \guidedvalue{54.36} & \guidedvalue{6.78} & \guidedvalue{48.99} & \guidedvalue{5.87} \\
\bottomrule
\end{tabular*}
\caption{Zero-shot Probe C gate pass rates (\%) on the full test set.
The opinion sentence is supplied.
All phrases must reach cosine 0.5, mean IoU must reach 0.5 over available boxes, and Emotion must match exactly.
The joint score requires all gates.
Bold marks the best result per column.}
\label{tab:guided_summary}
\end{table}

\paragraph{Region interpretation and localization give different rankings.}
Probe A supplies the reference region and asks for a role description.
InternVL scores above Qwen3.8-27B in phrase cosine, at 65.51 versus 64.27.
Their Probe B localization accuracies instead differ sharply, at 9.84\% versus 48.03\%.
MiniCPM has the highest zero-shot Probe A cosine, at 68.53, despite its low Probe C joint score of 5.87.
A model that describes a supplied region well may still struggle to select evidence for an opinion.
This distinction motivates testing both directions between phrases and regions.

\begin{figure}[t]
\centering
\includegraphics[width=\columnwidth]{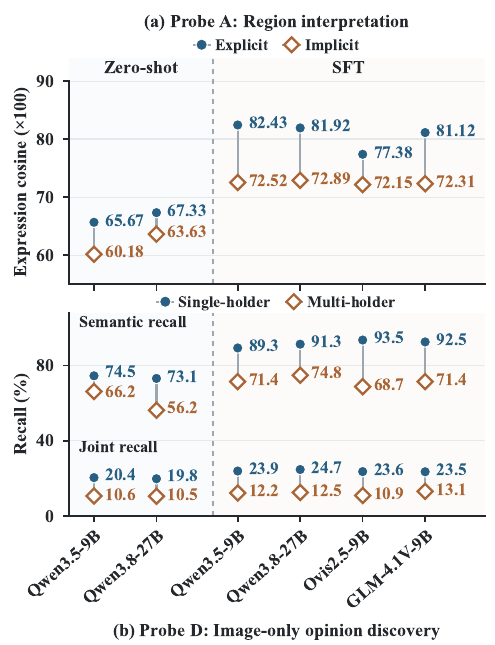}
\caption{Performance by opinion property.
(a) Probe A Expression cosine on 599 sampled opinions, split by explicit and implicit cues.
(b) Probe D semantic and joint recall in single-holder and multi-holder test scenes.
Lines connect group scores for each model.
Groups differ in scene content and opinion count.}
\label{fig:property_summary}
\end{figure}

\paragraph{Adaptation widens the implicit-cue gap.}
Figure~\ref{fig:property_summary}(a) compares Expression descriptions for explicit and implicit cues.
Both groups improve after Qwen adaptation, but explicit cues benefit more.
The gap grows from 5.49 to 9.91 points for Qwen3.5-9B and from 3.70 to 9.03 for Qwen3.8-27B.
These differences arise with the reference Expression box already supplied.
Implicit cues remain harder to describe within a supplied region.

\paragraph{Single-holder performance can hide missed perspectives.}
Figure~\ref{fig:property_summary}(b) compares scene groups that differ in content and opinion count.
Recall is lower in multi-holder scenes for every displayed model condition.
Among the four adapted models, Ovis has the highest single-holder semantic recall, at 93.5\%, but the lowest multi-holder recall, at 68.7\%.
Qwen3.8-27B has a smaller semantic gap, from 91.3\% to 74.8\%.
Its joint recall falls from 24.7\% to 12.5\% across the two scene groups.
Strong single-holder recall can mask missed attitudes across holders.

\subsection{RQ3: Effects of Adaptation}
\label{sec:adaptation}

\paragraph{Adaptation helps discovery more consistently than cue localization.}
All six adapted models improve Probe A cosine, Probe C joint score, and Probe D semantic F1.
Their paired confidence intervals lie above zero, as reported in Appendix~\ref{sec:appendix_full_results}.
Figure~\ref{fig:adaptation_summary} contrasts semantic discovery with Expression localization and joint opinion F1.
Joint F1 increases for all six models, with intervals above zero for five.
Qwen3.8-27B starts strongest and gains only 0.47 points, with an interval spanning zero.
InternVL starts lowest and gains 12.63 points, alongside a Holder accuracy increase from 9.50\% to 71.50\%.
GLM also improves output validity: its parse rate rises from 74.37\% to 99.51\% as semantic F1 rises by 49.02 points.
Under the same prompt, Qwen3.5-9B semantic F1 rises from 46.41 to 63.34 after SFT.
Expression accuracy changes from 15.23\% to 15.58\%.
Appendix~\ref{sec:adaptation_controls} gives the full comparison under the shared prompt.

\begin{figure}[t]
\centering
\includegraphics[width=\columnwidth]{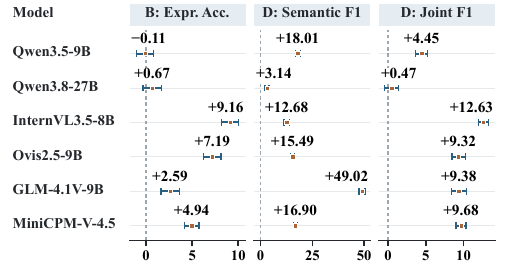}
\caption{Paired changes after SFT on 100 images, evaluated on the full test set.
Squares show SFT minus zero-shot in points, with 95\% image-bootstrap intervals.
Dashed lines mark zero change.
Panels use different horizontal scales.
Additional results appear in Table~\ref{tab:family_sft}.}
\label{fig:adaptation_summary}
\end{figure}

\paragraph{Better overall performance can accompany worse Target grounding.}
Four models have Probe B Expression gains with intervals above zero.
InternVL gains 9.16 points and Ovis2.5-9B gains 7.19, while both Qwen intervals span zero.
Even after adaptation, the best VLM reaches only 16.42\% Expression accuracy.
Fine-tuned Grounding DINO-B and Florence-2-L reach 14.95\% and 17.01\% respectively.
Meanwhile, Target accuracy drops by about six points for both Qwen models and GLM.
For Qwen3.8-27B, Holder accuracy rises from 81.37\% to 82.27\%, while Target accuracy falls from 47.72\% to 41.47\%.
The same model improves Holder localization while losing accuracy on attitude targets.

\paragraph{Filtered answers improve discovery.}
For Qwen3.5-9B, rejection-sampling fine-tuning (RFT) uses SFT-generated answers filtered against training annotations.
We retrain from the base model on these answers and the original targets.
Probe D filtering also caps the opinion count at the reference count.
Semantic F1 rises from 63.34 to 70.30 and joint F1 from 15.66 to 16.86.
Predicted opinions fall from 11,991 to 10,373.
Discovery improves, while Expression localization remains at 16.08\%.
Appendix~\ref{sec:alternatives} gives RFT details and comparisons with in-context demonstrations, training variants, and adapter merging.

\begin{table}[t]
\centering
\small
\setlength{\tabcolsep}{3pt}
\renewcommand{\arraystretch}{1.10}
\begin{tabular*}{\columnwidth}{@{\extracolsep{\fill}}lrr@{\hspace{8pt}}rr}
\toprule
 & \multicolumn{2}{c}{\textbf{Joint F1}} & \multicolumn{2}{c}{\textbf{Expr.\ gR}} \\
\cmidrule(lr){2-3}\cmidrule(lr){4-5}
\textbf{Text source} & VLM & \textbf{Snap} & VLM & \textbf{Snap} \\
\midrule
Qwen3.5-9B & 11.21 & 13.11 & 6.45 & 9.48 \\
Qwen3.8-27B & 15.50 & 17.02 & 8.71 & 9.54 \\
Qwen3.5-9B +SFT & 15.66 & 18.42 & 14.73 & 16.95 \\
Qwen3.5-9B +RFT & 16.86 & 18.81 & 14.97 & 16.97 \\
InternVL3.5-8B +SFT & 13.69 & 16.22 & 12.48 & 15.22 \\
\bottomrule
\end{tabular*}
\caption{Probe D with fixed opinion text and different boxes.
VLM uses the original boxes.
Snap selects the Grounding DINO proposal with highest IoU against each VLM box from the top 20 proposals.
Expr.\ gR denotes grounded Expression recall.}
\label{tab:reboxing_summary}
\end{table}

\paragraph{Discovery and grounding respond differently to data size.}
With 25 adaptation images, Qwen3.5-9B already reaches 78.01 in Probe A and 40.85 in Probe C.
The corresponding 100-image scores are 79.57 and 39.61.
Semantic discovery F1 instead increases from 46.27 to 53.33 and 63.34 with 25, 50, and 100 images.
Parse rate rises from 78.28\% to 84.82\% and 95.89\% over the same range.
Expression accuracy stays between 15.06\% and 16.22\%.
Its three 100-image seeds also cluster tightly, at 15.08\%, 15.58\%, and 16.20\%.
In these runs, more examples mainly improve discovery and output validity, while the cue-localization difficulty persists across seeds.
Appendix~\ref{sec:appendix_robustness} reports all runs.

\paragraph{Rough VLM boxes help select better detector proposals.}
We hold opinion text fixed and replace each box using Grounding DINO proposals.
Taking the highest-scoring detection changes joint F1 by $-1.67$ to $+0.37$ points across five text sources.
Selecting the proposal with highest overlap with the VLM box instead improves every source by 1.52--2.76 points.
All five paired intervals lie above zero.
This \emph{snap} rule uses the VLM box as a spatial hint, without reference boxes.
For the rejection-sampling model, joint F1 rises from 16.86 to 18.81 and grounded Expression recall from 14.97\% to 16.97\%.
The original boxes retain useful spatial information.
Refining their boundaries improves grounded opinion scores while preserving the descriptions.
Requiring IoU $\geq0.5$ with the original box reduces these gains.
For the two strongest text sources, the resulting Expression-recall intervals include zero.
A stricter overlap filter therefore discards some useful corrections.
Fine-tuning the detector changes joint F1 by at most 0.7 points for the top-1 and snap variants.
Proposal selection has a larger effect than detector adaptation in these comparisons.
Table~\ref{tab:reboxing_summary} summarizes the results, with all variants in Appendix~\ref{sec:appendix_pipeline}.

\FloatBarrier
\section{Conclusion}
\label{sec:conclusion}

We extend structured opinion mining to images through Visual Opinion Mining.
Each opinion links a holder, target, expression, and polarity to supporting visual evidence.
Our corpus and four probes separate region interpretation, evidence localization, guided grounding, and image-only discovery.
Expressive-cue grounding remains difficult even with phrases or opinion sentences supplied.
Adaptation improves discovery across six models, with uneven gains across roles and scenes.
Higher opinion scores can accompany poorer Target grounding.
Explicit cues benefit more than implicit cues in both Qwen models.
Multi-holder scenes remain difficult even for models with high single-holder recall.
Rough VLM boxes help select detector proposals that improve joint scores with opinion text fixed.
Expressive-cue grounding and multi-holder coverage remain distinct challenges for recovering structured opinions from images.

\section*{Limitations}

\paragraph{Annotations and visual evidence.}
The benchmark evaluates image-based interpretations against finite references.
Plausible opinions may be unannotated, so precision measures reference agreement (Section~\ref{sec:metrics}).
Unmatched predictions have not been manually adjudicated.
Annotations underwent review rather than independent double annotation (Appendix~\ref{sec:appendix_guidelines}).
Each grounded role has one reference box.
This representation approximates diffuse cues, and small Expression regions make IoU sensitive to boundary choices.

\paragraph{Scope of the comparisons.}
The study uses EMOTIC-derived images and tests adaptation on 100 multi-opinion scenes.
The findings concern this domain and supervision budget.
Property analyses use sampled, model-assisted labels and scene groups that differ in content and opinion count (Appendix~\ref{sec:appendix_statistics}).
The proprietary-model comparison covers 100 paired test images from an endpoint without pinned model snapshots (Appendix~\ref{sec:api}).
These comparisons describe performance under the reported sampling and inference conditions.

\paragraph{Evaluation conditions.}
Adaptation and test images are disjoint, but test data were used repeatedly during exploration, including coordinate pilots.
Pretraining exposure to source images is unknown.
Most training variants use one seed, and in-context runs use a fixed demonstration selection.
The adapted prompt permits whole-person boxes for cues too small to isolate.
GLM also switches to greedy answers without thinking, so its gains combine supervision and decoding changes.
Appendices~\ref{sec:appendix_reproducibility} and~\ref{sec:appendix_robustness} give the settings.

\section*{Ethics Statement}

\paragraph{Annotation and participant privacy.}
The six annotators received the research purpose and task requirements before giving informed consent.
They could withdraw from the study and were compensated at prevailing market rates.
Annotator records use anonymous identifiers.
The release will exclude names, contact details, and other personal information about annotators.
Appendix~\ref{sec:appendix_guidelines} details the annotation and review procedure.

\paragraph{Image access and intended use.}
EMOTIC permits use for non-commercial research and education.\footnote{\url{https://s3.sunai.uoc.edu/emotic/download.html}}
The source images remain subject to their applicable reuse conditions.
Release documentation will include source attribution and access conditions.
Where redistribution is restricted, we will provide permitted identifiers or derived annotations.
Our labels describe visually supported interpretations, rather than verified beliefs or mental states of depicted people.
Treating these labels as personal facts could reinforce stereotypes or support intrusive judgments.
The benchmark is intended for research on visual interpretation and grounding.
Dataset documentation will advise against surveillance, clinical assessment, and other high-stakes uses involving individuals.

\paragraph{Use of AI tools.}
The opinion quadruples and evidence boxes were annotated manually.
GPT-5.5 supplied auxiliary property labels, with sample-based manual checks described in Appendix~\ref{sec:appendix_statistics}.
Generative AI tools also assisted with grammar checking and polishing the manuscript during the writing process.

\bibliography{custom}
\appendix
\raggedbottom
\section{Annotation Guidelines and Quality Control}

\label{sec:appendix_guidelines}

\paragraph{Opinion unit.}
An opinion is an attitude that a visible person shows toward something in the image, supported by a visible cue.
For each opinion, the annotator writes one English sentence of the form ``The [Holder] [Expression] the [Target].''
The annotator then marks the three role phrases as spans of this sentence.
Each phrase must be a contiguous span that occurs verbatim in the sentence.
For ``The girl in the helmet smiles happily at the camera.'', the Target is ``the camera'', and a phrase such as ``camera/viewer'' is not allowed.
An image receives one opinion per supported Holder--Target pair, so several people in a scene, or one person reacting to several things, yield several opinions.
Annotators begin with the image and then describe the opinion.

\paragraph{What counts as an opinion.}
The image must show clear evidence that the Holder takes a subjective stance toward the Target.
This stance can be an emotion, evaluation, interest, attention, concern, comfort, blame, appreciation, fear, or surprise.
Helping, caring for, comforting, guiding, serving, or supporting another person counts, as in ``carefully supports the injured player''.
Assessing an object in an evaluative scene also counts, for example tasting wine, judging, inspecting damage, or watching a performance.
Focus on a screen, task, object, or person counts when it reflects attention, engagement, concern, or evaluation.
Purely physical actions such as standing, walking, or holding an object do not count.
Plain scene descriptions such as ``the man is at the table'' do not count either.
A phrase like ``looking at'' that only states where a person faces is too weak to be an opinion.

\paragraph{Holder.}
The Holder is a person, or a group acting together, who is visible in the image.
The phrase is a short noun phrase such as ``nurse'' or ``woman''.
When the scene contains several similar people, the phrase must identify one of them uniquely by position or appearance, as in ``man on the left'' or ``the blond child directly behind the boy in maroon''.
A phrase such as ``the child behind'' is not accepted when several children stand behind.
The Holder box covers the full visible body of that person or group, or the visible part when only the face or upper body appears.

\paragraph{Target.}
The Target is what the attitude is directed at: another person, a group, an animal, an object, or a scene or event.
It is written and disambiguated in the same way as the Holder.
Its box covers the visible extent of the entity, and a group Target receives one box around its visible members.
A Target that the sentence supports but the image does not show, such as the camera, the photographer, or a person outside the frame, receives no box.
An event can be a Target when both the image and sentence support an attitude toward that event.

\paragraph{Expression.}
The Expression is the observable behavior that shows the attitude.
The phrase is a verb phrase that names the behavior and its manner and includes the preposition that links it to the Target, as in ``smiles warmly at'', ``looks worried about'', or ``carefully examines''.
The Expression box marks the region that carries the cue.
This region is the face for a facial expression, and the face, eyes, and head orientation for gaze or attention.
It is the hands and arms for a gesture, and the contact or action region for helping, caring, or comforting.
It is the whole person when the cue is posture or body language.
When two people jointly form the evidence, the box may cover both.
The Expression box may overlap the Holder or Target box.
In the test set, it coincides with the Holder box for 1,605 of the 7,150 opinions (22.4\%).

\paragraph{Emotion.}
The emotion label is the polarity of the attitude of the Holder toward the Target as shown by the Expression.
It is Positive or Negative when the cue has a clear valence and Neutral otherwise, for example for plain attention.

\paragraph{Boxes.}
Each grounded role receives one axis-aligned box drawn tightly around the region described above.
Coordinates $[x_1,y_1,x_2,y_2]$ are normalized by image dimensions.

\paragraph{Status labels.}
Annotators mark each opinion as valid, invalid, or unsure.
Invalid opinions lack visual evidence or have a wrong, vague, or unsupported role and are removed.
An unsure opinion is plausible but uncertain, and the annotator records the reason in a note.
The test set retains 319 unsure opinions.
None of them has a Target box, so they enter text-based scores but not Target localization.

\paragraph{Review.}
Four annotators labeled the first 3,000 images.
Two reviewers checked this batch, each covering a disjoint part.
An opinion and its annotation were accepted only when the reviewer agreed with the sentence, role phrases, emotion, and boxes.
Otherwise the reviewer returned the opinion to its original annotator for revision.
Reviewers accepted 85\% of images on first review.
The two reviewers then labeled the remaining 2,239 images.
The reviewers then cross-checked annotations on randomly sampled images from this batch under the same criterion.
The reported first-review acceptance rate is about 80\% (159/200).
The annotations of the rejected images were revised.

\paragraph{Review unit and box granularity.}
Image-level acceptance covers opinion content, role phrases, emotion, and boxes together.
Box review follows the rules above: facial cues use the face, and posture may use the whole person.

\paragraph{Automatic checks.}
A validation script checks every record before evaluation.
Each box must have four coordinates with $x_1<x_2$ and $y_1<y_2$, and every opinion must have Holder and Expression boxes.
We map 39 free-form emotion labels, such as ``sadness'' and ``concerned'', to their polarity.
Seven records without an emotion label or a role phrase are removed.

Table~\ref{tab:annotation_examples} provides annotation examples.

\begin{table}[t]
\centering
\small
\setlength{\tabcolsep}{3pt}
\begin{tabular}{lllc}
\toprule
\textbf{Holder} & \textbf{Expression} & \textbf{Target} & \textbf{Emo.} \\
\midrule
priest & comforts & woman & Pos \\
woman & looks worried about & man & Neg \\
speaker & shows concern for & audience & Pos \\
man & looks solemnly at & crucifix & Neg \\
man & stands attentively at & Buddha statues & Neu \\
man & smiles warmly at & wheat field & Pos \\
\bottomrule
\end{tabular}
\caption{Annotated opinions with person, group, object, and scene Targets.}
\label{tab:annotation_examples}
\end{table}

\section{Dataset Statistics and Property Labels}

\label{sec:appendix_statistics}

\begin{table*}[tp]
\centering
\small
\setlength{\tabcolsep}{3pt}
\begin{tabular}{l l l l l}
\toprule
\textbf{System} & \textbf{Family} & \textbf{Size} & \textbf{Setting} & \textbf{Probes} \\
\midrule
\multicolumn{5}{l}{\textit{Open vision-language models (full test set)}} \\
Qwen3.5-9B           & Qwen      & 9B  & zero-shot, SFT & A--D \\
Qwen3.8-27B          & Qwen      & 27B & zero-shot, SFT & A--D \\
InternVL3.5-8B       & InternVL  & 8B  & zero-shot, SFT & A--D \\
Ovis2.5-9B           & Ovis      & 9B  & zero-shot, SFT & A--D \\
Ovis2.5-2B           & Ovis      & 2B  & zero-shot      & A--D \\
GLM-4.1V-9B-Thinking & GLM-V     & 9B  & zero-shot, SFT & A--D \\
MiniCPM-V-4.5        & MiniCPM-V & 8B  & zero-shot, SFT & A--D \\
\midrule
\multicolumn{5}{l}{\textit{Single-task vision models (full test set)}} \\
Grounding DINO-B     & DINO      & base & zero-shot, FT & B \\
OWLv2-B              & OWL       & base & zero-shot     & B \\
Florence-2-L         & Florence  & 0.8B & zero-shot, FT & A, B \\
\midrule
\multicolumn{5}{l}{\textit{Proprietary models (100-image subset)}} \\
GPT-5.5              & GPT       & --  & zero-shot      & A--D \\
Gemini-3.6-Flash     & Gemini    & --  & zero-shot      & A--D \\
\bottomrule
\end{tabular}
\caption{Evaluated systems.
SFT/FT: adapted on the 100 adaptation images.}
\label{tab:systems}
\end{table*}

The test partition averages 1.39 opinions per image and contains as many as 12 opinions for one image.
Of the 5,139 test images, 3,812 contain one opinion, 905 contain two, and 422 contain three or more.
Thus 1,327 images (25.8\%) have multiple annotated opinions.
Emotion labels are 64.3\% Positive, 22.5\% Negative, and 13.1\% Neutral.

\paragraph{Opinion properties.}
We identify multi-holder scenes using distinct annotated holders.
This label groups images by the number of distinct opinion holders, regardless of attitude polarity.
Multi-holder scenes account for 643 of the 5,139 test images (12.5\%).
They contain 1,807 of the 7,150 opinions (25.3\%).
Implicitness and compositionality are labeled by GPT-5.5 on a random sample of 599 opinions from 375 images, given the image, the annotated boxes, and a close-up of the Expression region.
An opinion is implicit when no readable facial expression shows the attitude and it has to be inferred from posture, gaze, action, or context.
In this sample, 76.0\% of opinions are implicit and 4.5\% are compositional.
A manual check of 200 labels yields 79.0\% agreement on implicitness and 96.5\% on compositionality.

\begin{figure}[t]
  \centering
  \includegraphics[width=\columnwidth]{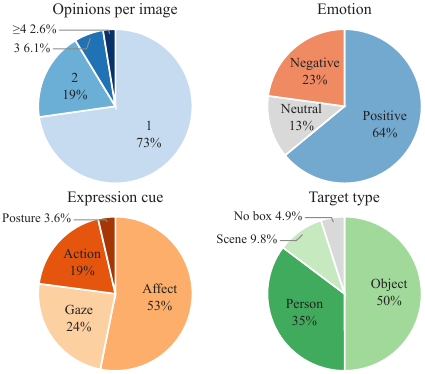}
  \caption{Composition of the full corpus of 5,239 images and 7,406 opinions.
  Opinion counts use images as the denominator, and the other panels use opinions.
  Expression-cue and Target-type categories are approximated by keyword rules over the annotated phrases.
 ``Scene'' includes events, and ``No box'' marks Targets without a reference box.}
  \label{fig:data_composition}
\end{figure}

Figure~\ref{fig:data_composition} summarizes the full corpus, including the adaptation partition.
Phrase-based categories differ from the model-assisted labels.

\section{Implementation and Reproducibility}
\label{sec:appendix_reproducibility}

Table~\ref{tab:systems} lists the evaluated systems and applicable probes.
The single-task detectors return their top-scoring box for the role phrase in Probe B.
Florence-2-L uses its phrase-grounding and region-description tasks.
Its zero-shot region prompt does not specify the opinion role.

\paragraph{Adaptation.}

We adapt six open models on the 100-image partition to test learning from limited supervision (Section~\ref{sec:split}).
Ovis2.5-2B is evaluated only zero-shot.
The 256 opinions are converted into 1,868 supervision instances covering all four probe formats.
All six models use LoRA on language-model attention and MLP projections.
The rank is 16, $\alpha=32$, and dropout is 0.05.
Training uses three epochs, learning rate $10^{-4}$, and effective batch size 16.
With no held-out data, the final checkpoint is used.
Adapted and zero-shot prompts use the same query content and output schema.
The adapted instruction additionally permits a whole-person Expression box when the cue is too small to isolate.
Training targets contain only the structured answer.
All adapted models use greedy decoding, including GLM-4.1V-9B-Thinking without a thinking trace.
Grounding DINO-B is fine-tuned on Probe B queries from the same 100 images.
It uses three epochs, learning rate $10^{-5}$, backbone learning rate $10^{-6}$, and a frozen text encoder.
We use the final checkpoint.
Florence-2-L is fully fine-tuned on their Probe A and Probe B queries for three epochs.
The learning rate is $10^{-5}$, with $10^{-6}$ for the vision encoder.
Its region prompt additionally names the opinion role.
Adaptation uses only the designated 100-image partition.

\paragraph{Prompts, decoding, and coordinates.}

Each model receives a task-specific prompt requesting structured JSON.
Appendix~\ref{sec:appendix_prompts} lists the full prompts.
Boxes use $[x_1,y_1,x_2,y_2]$ on the registered coordinate grid of each model and are converted to a common percentage representation before scoring.
The coordinate convention of each open model was selected or verified with a small coordinate pilot before its full run.
Open models are decoded greedily with thinking disabled, except zero-shot GLM-4.1V-9B-Thinking, which uses its recommended sampling settings with thinking enabled.

\paragraph{Metric details.}
Appendix~\ref{sec:appendix_metrics} defines parsing, opinion matching, and scoring for all four probes.

\paragraph{Repeated runs.}
Qwen3.5-9B adaptation on 100 images uses three seeds.
The main tables use seed 42.
The 25- and 50-image runs, other five adapted VLMs, training variants, and adapted specialists use one seed.

\paragraph{Hardware and software.}
All open-model experiments run on NVIDIA H100 80GB GPUs, using up to four GPUs per job.
Open VLMs are served with vLLM~0.26.0.
Adaptation uses PyTorch~2.13, Transformers~5.14, and PEFT~0.20.

\section{Prompts and Input Formats}
\label{sec:appendix_prompts}

\newcommand{\ph}[1]{\textnormal{\textit{$\langle$#1$\rangle$}}}
\newcommand{\promptbox}[2]{\par\medskip\begingroup\small\raggedright\noindent\textbf{#1.}\enspace #2\par\endgroup\medskip}

All vision-language models, open and proprietary, receive the prompts below.
Each query consists of a system message and a user message that contains the instruction text followed by the image.
All models use a 0--1000 coordinate grid.
The placeholder \ph{coord} reads ``0-1000 normalized to image size (top-left = 0,0; bottom-right = 1000,1000)'', and \ph{unit} in Probe A reads ``0-1000 (normalized grid)''.
\ph{role} is holder, target, or expression, and \ph{role description} is the matching line below.

\promptbox{Role descriptions}{%
holder: the entity (person or group) that expresses the opinion\\
target: what the opinion is about (an entity, object, or event)\\
expression: the visual or behavioral cue that conveys the opinion}

\promptbox{Probe A, system}{%
You are a visual opinion mining system. A bounding box marks one element of an opinion in the image. Describe the role that region plays IN THAT OPINION -- an opinion-aware phrase, NOT a generic object name. Do NOT answer with a plain label like 'man's face', 'hand', or 'table'. Instead:\\
\hspace*{1em}- holder: who expresses the opinion (e.g. 'the smiling woman')\\
\hspace*{1em}- target: what the opinion is about (e.g. 'the colorful candies')\\
\hspace*{1em}- expression: the VISIBLE CUE that conveys the attitude -- a behavior, posture, facial expression, gaze, gesture, or spatial relation (e.g. 'smiling at the child', 'leaning away from the man', 'focused gaze toward the laptop') -- NOT an object and NOT an abstract emotion word.\\
Answer with a short phrase, not a full sentence.}

\promptbox{Probe A, user}{%
A bounding box [\ph{box}] (coordinates in \ph{unit}) marks the \ph{role} of an opinion in this image (\ph{role description}).\ph{extra} In a few words, give the opinion-aware phrase for this \ph{role} -- not a plain object name. Respond with ONLY JSON: \{"phrase": "..."\}.\\[3pt]
\ph{extra}, for the expression role only:\\
\hspace*{1em}Describe the visible cue (behavior/posture/facial expression/gaze/spatial relation) it shows, not the object itself.}

\promptbox{Probe B, system}{%
You localize a described element in an image with a single bounding box [x1,y1,x2,y2] in coordinates \ph{coord}. Box tightly around the described element only. For an expression (a behavioral/visual cue), box the visible cue that conveys the opinion -- the face, gaze, hands, gesture, or posture that shows it -- NOT the whole person, unless the whole-body posture itself is the cue.}

\promptbox{Probe B, user}{%
In this image, give the bounding box of the \ph{role} described as: "\ph{phrase}" (this is \ph{role description}).\ph{extra} Respond with ONLY JSON: \{"box": [x1,y1,x2,y2]\} with coordinates \ph{coord}.\\[3pt]
\ph{extra}, for the expression role only:\\
\hspace*{1em}Box the visible cue (face/gaze/hands/posture) that conveys it, not the whole person unless the whole-body posture is the cue.}

\promptbox{Probe C, system}{%
You are a visual opinion mining system. Given an image and a sentence that states one opinion present in the scene, extract that opinion as (holder, expression, target, emotion) and localize holder/expression/target with bounding boxes [x1,y1,x2,y2] in coordinates \ph{coord}. expression is the VISIBLE CUE in the image that conveys the opinion -- a behavior, posture, facial expression, gaze, or gesture (e.g. 'smiling face', 'leaning away') -- NOT an abstract emotion word like 'happy' or 'angry'. Draw the expression box TIGHTLY on the visible cue (the face/gaze/hands/gesture/posture that shows it), NOT around the whole holder -- unless the whole-body posture itself is the cue. emotion is one of Positive, Negative, Neutral. If the target is not visible, set its box to null.}

\promptbox{Probe C, user}{%
Opinion sentence: "\ph{sentence}"\\
Extract and localize this opinion. Respond with ONLY JSON, no prose:\\
\{"holder": str, "holder\_box": [x1,y1,x2,y2], "expression": str, "expression\_box": [x1,y1,x2,y2], "target": str, "target\_box": [x1,y1,x2,y2] or null, "emotion": "Positive"|"Negative"|"Neutral"\}\\
Coordinates are \ph{coord}.}

\promptbox{Probe D, system}{%
You are a visual opinion mining system. Given an image, you extract every distinct OPINION expressed in the scene. An opinion has four parts:\\
\hspace*{1em}- holder: the entity (person/group) expressing the opinion\\
\hspace*{1em}- expression: the visible/behavioral cue that conveys the opinion -- a posture, facial expression, gaze, gesture, or spatial relation (a short phrase), NOT an abstract emotion word\\
\hspace*{1em}- target: what the opinion is about (may be another entity, an object, or an event)\\
\hspace*{1em}- emotion: one of Positive, Negative, Neutral\\
For holder, expression, and target you also give a bounding box as [x1,y1,x2,y2] in coordinates \ph{coord}. If the target is not visible in the image, set its box to null. Draw the expression box TIGHTLY on the visible cue (the face/gaze/hands/gesture/posture that shows it), NOT around the whole holder -- unless the whole-body posture itself is the cue.\\
What counts as an opinion:\\
\hspace*{1em}- Do NOT output purely objective actions or scene descriptions (e.g. 'a man is holding a cup', 'a woman is walking').\\
\hspace*{1em}- Only output an opinion when there is VISIBLE evidence of a subjective stance: an attitude, affect, preference, attention, engagement, discomfort, approval, or similar.\\
\hspace*{1em}- Every opinion you output must be grounded by visible evidence in the image.}

\promptbox{Probe D, user}{%
Extract all opinions from this image. Skip purely objective actions or scene descriptions -- include an opinion only when a visible cue shows a subjective stance. Respond with ONLY a JSON array, no prose. Each element:\\
\{"holder": str, "holder\_box": [x1,y1,x2,y2], "expression": str, "expression\_box": [x1,y1,x2,y2], "target": str, "target\_box": [x1,y1,x2,y2] or null, "emotion": "Positive"|"Negative"|"Neutral"\}\\
Coordinates are \ph{coord}. Output [] if there are no opinions.}

\paragraph{Adapted prompts.}
Both prompt versions permit whole-person boxes when posture carries the cue.
The adapted version also permits them when the person is too small to isolate the cue.
The excerpts below show the original and revised instructions.

\promptbox{Probes C and D, system}{%
Zero-shot: Draw the expression box TIGHTLY on the visible cue (the face/gaze/hands/gesture/posture that shows it), NOT around the whole holder -- unless the whole-body posture itself is the cue.\\[3pt]
Adapted: Draw the expression box on the visible cue that conveys the opinion (the face/gaze/hands/gesture/posture that shows it). If the holder is so small that the cue cannot be boxed on its own, or the cue is the whole-body posture, the expression box may be the whole holder.}

\promptbox{Probe B, system}{%
Zero-shot: \dots box the visible cue that conveys the opinion -- the face, gaze, hands, gesture, or posture that shows it -- NOT the whole person, unless the whole-body posture itself is the cue.\\[3pt]
Adapted: \dots box the visible cue that conveys the opinion -- the face, gaze, hands, gesture, or posture that shows it. If the person is so small that the cue cannot be boxed on its own, or the cue is the whole-body posture, box the whole person.}

\promptbox{Probe B, user, expression role}{%
Zero-shot: Box the visible cue (face/gaze/hands/posture) that conveys it, not the whole person unless the whole-body posture is the cue.\\[3pt]
Adapted: Box the visible cue (face/gaze/hands/posture) that conveys it; box the whole person if the person is too small for the cue to be boxed on its own or the whole-body posture is the cue.}

\paragraph{In-context demonstrations.}
In-context runs use the adapted prompts.
Each demonstration is a user turn with the instruction and image of an adaptation item, followed by an assistant turn with its reference answer in the required JSON format.
The $k$ demonstrations follow the system message and precede the query.
A Probe D demonstration covers one image with all its opinions, a Probe C demonstration covers one opinion, and a Probe A or B demonstration covers one role query.
Demonstration boxes use the coordinate grid of the model.

\paragraph{Single-task models.}
Grounding DINO-B and OWLv2-B receive the role phrase as the text query, and OWLv2 truncates it to its 16-token limit.
Florence-2-L uses its phrase-grounding task with the role phrase for Probe B and reads the first returned box.
For zero-shot Probe A it uses its region-description task.
After fine-tuning, Probe A names the role: ``What is the opinion \ph{role} in the region \ph{box}?''
The box uses Florence location tokens.

\section{Evaluation Metrics}
\label{sec:appendix_metrics}

\paragraph{Output parsing.}
The scorer extracts the first JSON object or array from each output, after removing any thinking trace and Markdown code fences.
An output that yields no valid JSON counts as an empty prediction, and parse rate is the share of queries whose output parses.
Predicted boxes are converted from the 0--1000 grid to percentages of image width and height.
Corners are reordered so that $x_1\leq x_2$ and $y_1\leq y_2$, and coordinates are clipped to the image.
Gemini boxes are read in $y$-first order (Appendix~\ref{sec:api}).

\paragraph{Phrase similarity and box overlap.}
Phrase similarity $\operatorname{sim}(\hat p,p)$ is the cosine between the L2-normalized embeddings of the predicted phrase $\hat p$ and the reference phrase $p$ from BGE-small-en-v1.5~\citep{xiao2023cpack}.
A missing, empty, or non-string prediction receives similarity 0.
Box overlap is the intersection over union $\operatorname{IoU}(\hat{\mathbf b},\mathbf b)$, and a missing or invalid predicted box receives IoU 0.
A phrase passes when $\operatorname{sim}\geq0.5$, and a box passes when $\operatorname{IoU}\geq0.5$.

\paragraph{Probe A.}
Each query supplies one reference box and its role, and the model returns one phrase.
The score is the mean similarity over all 21,096 queries.
We also report the mean for each role.

\paragraph{Probe B.}
Each query supplies one role phrase, and the model returns one box.
Accuracy is the share of queries whose box passes.
We report it for each role and over all 21,096 queries.

\paragraph{Probe C.}
Each of the 7,150 queries supplies one opinion sentence, and the model returns one quadruple with boxes.
A prediction passes three gates.
The semantic gate requires all three role phrases to pass.
The emotion gate requires the exact reference label.
The spatial gate requires the mean IoU over the roles $R(o)$ with a reference box to reach 0.5,
\begin{equation}
g(\hat o,o)=\frac{1}{|R(o)|}\sum_{r\in R(o)}\operatorname{IoU}(\hat{\mathbf b}^{r},\mathbf b^{r})\geq0.5 .
\end{equation}
The joint score is the share of queries that pass all three gates.
Each query has exactly one prediction, so this share equals precision, recall, and F1.
We also report the pass rate of each gate, the semantic pass rate of each role, and per-role box accuracy over the opinions that have that reference box.

\paragraph{Probe D matching.}
Probe D compares the predicted opinions $\hat O$ and reference opinions $O$ of each image.
Every predicted--reference pair within the image is a candidate when it satisfies the criterion of the metric, and each candidate receives a score.
Candidates are sorted by score in descending order.
A candidate is accepted when neither its prediction nor its reference has been matched already.
This greedy procedure yields a one-to-one matching for each image.
Counts are summed over all 5,139 test images, so precision is $\mathrm{TP}/|\hat O|$, recall is $\mathrm{TP}/|O|$, and F1 is their harmonic mean.

\paragraph{Probe D scores.}
For semantic discovery, a pair is a candidate when all three role phrases pass, and its score is the mean of the three similarities.
Semantic F1 follows from this matching.
The three phrases describe different parts of the opinion: participant identity, the object of the attitude, and the behavior expressing it.
All three similarities must pass the threshold for the pair to enter semantic matching.
Emotion is evaluated separately from these phrase checks.
Emotion accuracy is the share of these semantic matches whose predicted label equals the reference.
For joint opinion F1, a pair is a candidate when all three phrases pass, the emotion is exact, and $g(\hat o,o)\geq0.5$.
Its score is the mean similarity plus the mean IoU, and joint matching runs separately from semantic matching.
The spatial criterion averages over boxes, so a joint match can contain one role box below 0.5.

\paragraph{Probe D role-level scores.}
Each role is matched on its own.
For role $r$, a pair is a candidate when the role phrase passes, and its score is the role similarity, with IoU breaking ties.
Role precision, recall, and F1 follow from this matching.
Conditional box accuracy is the share of role matches with a reference box whose predicted box passes.
Grounded recall divides the number of role matches with a passing box by the number of reference boxes of that role, including those of undiscovered opinions.
Grounded recall therefore penalizes both missed discovery and poor localization.

\paragraph{Bootstrap intervals.}
Paired differences use an image-level bootstrap with 2,000 resamples.
Each resample draws test images with replacement and keeps all opinions and queries of a drawn image together.
All compared systems are scored on the same resampled images.
Intervals span the 2.5th to 97.5th percentiles of the resampled differences.

\section{Full Diagnostic Results}
\label{sec:appendix_full_results}

Tables~\ref{tab:probe_a}--\ref{tab:probe_d} report the full results for Probes A--D.
The role-level scores separate phrase recovery from localization, and Probe D further separates conditional accuracy from grounded recall.
Proprietary-model rows use the 100-image subset.
Table~\ref{tab:api100} provides the corresponding open-model scores for paired comparisons.
Table~\ref{tab:family_sft} reports paired adaptation effects for every summary metric.

\begin{table*}[tp]
\begin{minipage}[t]{0.48\textwidth}
\centering
\footnotesize
\setlength{\tabcolsep}{2pt}
\begin{tabular}{l cccc}
\toprule
\textbf{Model} & Holder & Target & Expr. & All \\
\midrule
\multicolumn{5}{l}{\textit{Single-task region-description model}} \\
Florence-2-L                 & 64.68 & 66.84 & 58.60 & 63.31 \\
Florence-2-L~+FT             & \textbf{83.22} & \textbf{75.86} & \textbf{72.43} & \textbf{77.19} \\
\midrule
\multicolumn{5}{l}{\textit{Zero-shot vision-language models}} \\
Qwen3.5-9B                   & 67.33 & 67.15 & 61.74 & 65.38 \\
Qwen3.8-27B                  & 62.69 & 65.54 & 64.64 & 64.27 \\
InternVL3.5-8B               & 65.01 & 65.35 & 66.16 & 65.51 \\
Ovis2.5-9B                   & 66.51 & 68.44 & 65.94 & 66.94 \\
Ovis2.5-2B                   & 66.35 & 64.58 & 64.82 & 65.26 \\
GLM-4.1V-9B         & 70.54 & 67.95 & \textbf{66.58} & 68.37 \\
MiniCPM-V-4.5                & \textbf{70.66} & \textbf{69.26} & 65.72 & \textbf{68.53} \\
\midrule
\multicolumn{5}{l}{\textit{Adapted on the 100 adaptation images (LoRA SFT)}} \\
Qwen3.5-9B~+SFT              & 84.84 & 78.24 & 75.56 & 79.57 \\
Qwen3.8-27B~+SFT             & \textbf{85.23} & \textbf{78.85} & \textbf{75.82} & \textbf{79.99} \\
InternVL3.5-8B~+SFT          & 82.54 & 74.34 & 74.58 & 77.20 \\
Ovis2.5-9B~+SFT              & 84.47 & 76.35 & 73.35 & 78.09 \\
GLM-4.1V-9B~+SFT    & 85.01 & 77.49 & 74.90 & 79.16 \\
MiniCPM-V-4.5~+SFT           & 85.11 & 78.57 & 74.69 & 79.47 \\
\midrule
\multicolumn{5}{l}{\textit{Proprietary, zero-shot (100-image subset only)$^\dagger$}} \\
GPT-5.5$^\dagger$            & 63.78 & 65.68 & 67.63 & 65.70 \\
Gemini-3.6-Flash$^\dagger$   & 61.56 & 62.74 & 65.14 & 63.15 \\
\bottomrule
\end{tabular}
\caption{Probe~A (box$\to$phrase): mean BGE cosine ($\times100$) by role.
Bold marks the best result per full-test block.
$^\dagger$Rows use only the 100-image test subset and are not directly comparable with full-test rows.}
\label{tab:probe_a}
\end{minipage}\hfill
\begin{minipage}[t]{0.48\textwidth}
\centering
\footnotesize
\setlength{\tabcolsep}{2pt}
\begin{tabular}{l cccc}
\toprule
\textbf{System} & Holder & Target & Expr. & All \\
\midrule
\multicolumn{5}{l}{\textit{Spatial prior}} \\
Whole image                  & 12.73 & 9.05 & 1.22 & 7.64 \\
\midrule
\multicolumn{5}{l}{\textit{Single-task grounding models}} \\
Grounding DINO-B             & 76.08 & \textbf{42.42} & 14.03 & 44.21 \\
OWLv2-B                      & 58.94 & 36.73 & 4.91 & 33.47 \\
Florence-2-L                 & 73.52 & 40.08 & 11.10 & 41.60 \\
Grounding DINO-B~+FT         & 77.89 & 42.25 & 14.95 & 45.07 \\
Florence-2-L~+FT             & \textbf{78.90} & 40.98 & \textbf{17.01} & \textbf{45.71} \\
\midrule
\multicolumn{5}{l}{\textit{Zero-shot vision-language models}} \\
Qwen3.5-9B                   & 80.71 & 46.47 & \textbf{15.69} & 47.64 \\
Qwen3.8-27B                  & \textbf{81.37} & \textbf{47.72} & 14.98 & \textbf{48.03} \\
InternVL3.5-8B               & 9.50 & 17.13 & 3.24 & 9.84 \\
Ovis2.5-9B                   & 65.58 & 30.43 & 4.84 & 33.67 \\
Ovis2.5-2B                   & 64.08 & 33.65 & 8.62 & 35.48 \\
GLM-4.1V-9B         & 79.57 & 47.37 & 13.83 & 46.91 \\
MiniCPM-V-4.5                & 75.30 & 38.68 & 7.02 & 40.36 \\
\midrule
\multicolumn{5}{l}{\textit{Adapted on the 100 adaptation images (LoRA SFT)}} \\
Qwen3.5-9B~+SFT              & 81.76 & 40.14 & 15.58 & 45.92 \\
Qwen3.8-27B~+SFT             & \textbf{82.27} & \textbf{41.47} & 15.65 & \textbf{46.54} \\
InternVL3.5-8B~+SFT          & 71.50 & 32.90 & 12.41 & 39.04 \\
Ovis2.5-9B~+SFT              & 78.52 & 36.89 & 12.03 & 42.57 \\
GLM-4.1V-9B~+SFT    & 81.40 & 41.32 & \textbf{16.42} & 46.46 \\
MiniCPM-V-4.5~+SFT           & 79.23 & 36.60 & 11.96 & 42.70 \\
\midrule
\multicolumn{5}{l}{\textit{Proprietary, zero-shot (100-image subset only)$^\dagger$}} \\
GPT-5.5$^\dagger$            & 74.38 & 29.03 & 8.75 & 37.47 \\
Gemini-3.6-Flash$^\dagger$   & 68.13 & 30.97 & 10.00 & 36.42 \\
\bottomrule
\end{tabular}
\caption{Probe~B (phrase$\to$box): Acc@IoU$_{0.5}$ by role.
Bold marks the best result per full-test block.
$^\dagger$Rows use only the 100-image test subset and are not directly comparable with full-test rows.}
\label{tab:probe_b}
\end{minipage}
\end{table*}

\begin{table*}[tp]
\centering
\scriptsize
\setlength{\tabcolsep}{3.4pt}
\resizebox{\textwidth}{!}{%
\begin{tabular}{l cccc cccc c c}
\toprule
& \multicolumn{4}{c}{\textbf{Phrase} (BGE $\geq0.5$)} & \multicolumn{4}{c}{\textbf{Box} (IoU $\geq0.5$)} & & \\
\cmidrule(lr){2-5} \cmidrule(lr){6-9}
\textbf{Model} & H & T & E & All three & H & T & E & Mean-IoU gate & Emotion & Joint score \\
\midrule
\multicolumn{11}{l}{\textit{Zero-shot vision-language models}} \\
Qwen3.5-9B                   & \textbf{99.94} & 96.29 & 98.18 & 94.50 & \textbf{82.57} & 43.00 & 15.40 & 42.38 & 78.08 & 31.50 \\
Qwen3.8-27B                  & \textbf{99.94} & \textbf{99.66} & 99.51 & \textbf{99.15} & 82.15 & \textbf{47.40} & \textbf{17.55} & \textbf{48.01} & 79.06 & \textbf{38.29} \\
InternVL3.5-8B               & 99.83 & 96.94 & 99.52 & 96.48 & 14.69 & 17.03 & 4.31 & 4.92 & 83.69 & 3.94 \\
Ovis2.5-9B                   & 99.87 & 98.27 & \textbf{99.54} & 97.80 & 73.07 & 36.07 & 11.26 & 26.10 & 80.64 & 20.99 \\
Ovis2.5-2B                   & 99.76 & 99.09 & 99.03 & 98.22 & 59.52 & 31.31 & 8.54 & 18.18 & 84.83 & 15.51 \\
GLM-4.1V-9B-Thinking         & 99.73 & 98.62 & \textbf{99.54} & 98.28 & 79.69 & 44.44 & 13.29 & 39.20 & \textbf{86.99} & 33.90 \\
MiniCPM-V-4.5                & 55.58 & 54.76 & 55.23 & 54.36 & 51.71 & 23.34 & 6.92 & 6.78 & 48.99 & 5.87 \\
\midrule
\multicolumn{11}{l}{\textit{Adapted on the 100 adaptation images (LoRA SFT)}} \\
Qwen3.5-9B~+SFT              & \textbf{100.00} & 99.62 & \textbf{100.00} & 99.62 & \textbf{83.89} & 41.58 & 20.70 & 44.10 & \textbf{89.68} & 39.61 \\
Qwen3.8-27B~+SFT             & 99.97 & \textbf{99.92} & 99.97 & \textbf{99.92} & 83.39 & \textbf{42.05} & 20.62 & \textbf{44.48} & 89.48 & \textbf{39.64} \\
InternVL3.5-8B~+SFT          & 99.99 & 99.65 & 99.83 & 99.47 & 80.35 & 37.49 & 18.67 & 39.58 & 88.46 & 34.80 \\
Ovis2.5-9B~+SFT              & \textbf{100.00} & 99.68 & 99.99 & 99.66 & 80.91 & 38.92 & 16.43 & 38.50 & 88.55 & 34.01 \\
GLM-4.1V-9B-Thinking~+SFT    & 99.99 & 99.86 & 99.99 & 99.85 & 82.63 & 40.07 & \textbf{20.78} & 43.96 & 89.23 & 39.05 \\
MiniCPM-V-4.5~+SFT           & 99.94 & 99.75 & 99.90 & 99.71 & 80.38 & 34.23 & 16.81 & 32.17 & 89.23 & 28.69 \\
\midrule
\multicolumn{11}{l}{\textit{Proprietary, zero-shot (100-image subset only)$^\dagger$}} \\
GPT-5.5$^\dagger$            & 100.00 & 100.00 & 100.00 & 100.00 & 73.75 & 29.68 & 11.25 & 28.12 & 76.25 & 20.00 \\
Gemini-3.6-Flash$^\dagger$   & 100.00 & 99.38 & 100.00 & 99.38 & 78.12 & 29.68 & 10.00 & 32.50 & 85.62 & 28.75 \\
\bottomrule
\end{tabular}}
\caption{Probe C with a supplied opinion sentence: phrase, box, Emotion, and joint pass rates (\%).
H, T, and E denote Holder, Target, and Expression.
The spatial gate uses mean IoU over available reference boxes, rather than requiring each box to pass.
The joint score requires all phrase thresholds, exact Emotion, and the spatial gate.
Bold marks the best result per full-test block.
$^\dagger$Rows use only the 100-image test subset and are not directly comparable with full-test rows.}
\label{tab:probe_c}
\end{table*}

\begin{table*}[!t]
\centering
\small
\textbf{(a) Semantic discovery, Emotion, and output diagnostics}\\[3pt]
\begin{tabular*}{\textwidth}{@{\extracolsep{\fill}}lrrrrrrr@{}}
\toprule
Model & H F1 & T F1 & E F1 & Quad.\ F1 & Emo.\ Acc & \#Pred & Parse \\
\midrule
\midrule
\multicolumn{8}{l}{\textit{Zero-shot vision-language models}} \\
Qwen3.5-9B & 48.92 & 46.52 & 49.24 & 45.33 & 60.87 & 15,689 & 98.72 \\
Qwen3.8-27B & 64.69 & \textbf{63.68} & 65.82 & \textbf{61.24} & 60.04 & 8,927 & 100.00 \\
InternVL3.5-8B & 62.56 & 48.48 & 62.63 & 47.55 & 41.89 & 12,950 & 99.49 \\
Ovis2.5-9B & \textbf{67.11} & 57.46 & \textbf{67.79} & 56.06 & 50.80 & 8,209 & 100.00 \\
Ovis2.5-2B & 61.28 & 59.62 & 61.59 & 58.18 & 67.45 & 13,717 & 99.81 \\
GLM-4.1V-9B & 22.43 & 20.39 & 22.52 & 19.93 & \textbf{87.91} & 1,481 & 74.37 \\
MiniCPM-V-4.5 & 57.10 & 54.60 & 57.72 & 53.35 & 56.91 & 14,963 & 98.77 \\
\midrule
\multicolumn{8}{l}{\textit{Adapted on the 100 adaptation images (LoRA SFT)}} \\
Qwen3.5-9B~+SFT & 65.09 & 64.43 & 65.53 & 63.34 & 62.16 & 11,991 & 95.89 \\
Qwen3.8-27B~+SFT & 66.18 & 65.72 & 66.70 & 64.38 & 63.49 & 12,198 & 99.65 \\
InternVL3.5-8B~+SFT & 62.15 & 61.16 & 62.34 & 60.23 & \textbf{63.75} & 12,753 & 94.53 \\
Ovis2.5-9B~+SFT & \textbf{73.43} & \textbf{72.50} & \textbf{73.92} & \textbf{71.55} & 61.99 & 10,279 & 99.98 \\
GLM-4.1V-9B~+SFT & 71.12 & 70.06 & 71.32 & 68.95 & 59.20 & 10,924 & 99.51 \\
MiniCPM-V-4.5~+SFT & 72.22 & 71.61 & 73.13 & 70.26 & 60.15 & 10,451 & 99.55 \\
\midrule
\multicolumn{8}{l}{\textit{Proprietary, zero-shot (100-image subset only)$^\dagger$}} \\
GPT-5.5$^\dagger$ & 61.25 & 58.00 & 61.72 & 57.08 & 56.10 & 271 & 100.00 \\
Gemini-3.6-Flash$^\dagger$ & 66.83 & 62.53 & 67.78 & 61.58 & 59.69 & 259 & 100.00 \\
\bottomrule
\end{tabular*}

\medskip
\textbf{(b) Spatial grounding and joint opinion scores}\\[3pt]
\begin{tabular*}{\textwidth}{@{\extracolsep{\fill}}lrrrrrrrrr@{}}
\toprule
Model & H Acc & T Acc & E Acc & H gR & T gR & E gR & Joint P & Joint R & Joint F1 \\
\midrule
\midrule
\multicolumn{10}{l}{\textit{Zero-shot vision-language models}} \\
Qwen3.5-9B & 67.51 & 34.99 & 8.20 & \textbf{52.74} & \textbf{26.04} & 6.45 & 8.16 & \textbf{17.90} & 11.21 \\
Qwen3.8-27B & 69.69 & 35.74 & \textbf{11.77} & 50.69 & 25.71 & \textbf{8.71} & 13.96 & 17.43 & \textbf{15.50} \\
InternVL3.5-8B & 19.55 & 14.48 & 2.80 & 17.19 & 9.96 & 2.46 & 0.82 & 1.48 & 1.05 \\
Ovis2.5-9B & 68.26 & 30.65 & 6.72 & 49.20 & 19.17 & 4.90 & 6.94 & 7.97 & 7.42 \\
Ovis2.5-2B & 48.12 & 21.50 & 5.87 & 43.03 & 18.78 & 5.27 & 4.39 & 8.42 & 5.77 \\
GLM-4.1V-9B & \textbf{76.03} & \textbf{43.17} & 11.21 & 10.29 & 5.44 & 1.52 & \textbf{20.86} & 4.32 & 7.16 \\
MiniCPM-V-4.5 & 39.06 & 14.74 & 3.87 & 34.49 & 12.51 & 3.45 & 1.82 & 3.80 & 2.46 \\
\midrule
\multicolumn{10}{l}{\textit{Adapted on the 100 adaptation images (LoRA SFT)}} \\
Qwen3.5-9B~+SFT & \textbf{76.45} & \textbf{33.15} & 16.79 & 66.60 & 28.69 & 14.73 & 12.50 & 20.97 & 15.66 \\
Qwen3.8-27B~+SFT & 74.37 & 32.45 & 16.60 & 66.59 & \textbf{28.93} & 14.98 & 12.67 & \textbf{21.61} & 15.97 \\
InternVL3.5-8B~+SFT & 76.35 & 32.83 & 14.38 & 66.04 & 28.09 & 12.48 & 10.68 & 19.05 & 13.69 \\
Ovis2.5-9B~+SFT & 73.34 & 29.53 & 14.27 & 65.64 & 26.15 & 12.85 & \textbf{14.19} & 20.41 & \textbf{16.74} \\
GLM-4.1V-9B~+SFT & 74.51 & 31.57 & \textbf{17.98} & \textbf{66.98} & 28.03 & \textbf{16.21} & 13.69 & 20.91 & 16.54 \\
MiniCPM-V-4.5~+SFT & 70.86 & 25.44 & 13.67 & 62.99 & 22.48 & 12.31 & 10.22 & 14.94 & 12.14 \\
\midrule
\multicolumn{10}{l}{\textit{Proprietary, zero-shot (100-image subset only)$^\dagger$}} \\
GPT-5.5$^\dagger$ & 59.09 & 13.93 & 4.51 & 48.75 & 10.97 & 3.75 & 4.80 & 8.13 & 6.03 \\
Gemini-3.6-Flash$^\dagger$ & 66.43 & 19.84 & 3.52 & 58.13 & 16.13 & 3.12 & 7.34 & 11.87 & 9.07 \\
\bottomrule
\end{tabular*}
\caption{Probe D results split into semantic and spatial panels.
Box accuracy is conditional on semantic matches, while grounded recall (gR) includes undiscovered elements.
Joint scores require role semantics, exact Emotion, and the mean-IoU gate.
Bold marks the best result per full-test block.
$^\dagger$Rows use only the 100-image subset and are not directly comparable with full-test rows.}
\label{tab:probe_d}
\end{table*}

\begin{table*}[tp]
\centering
\small
\setlength{\tabcolsep}{4pt}
\begin{tabular}{l cc c c cc}
\toprule
& \textbf{Probe A} & \multicolumn{2}{c}{\textbf{Probe B}} & \textbf{Probe C} & \multicolumn{2}{c}{\textbf{Probe D}} \\
\cmidrule(lr){2-2} \cmidrule(lr){3-4} \cmidrule(lr){5-5} \cmidrule(lr){6-7}
\textbf{Model (SFT $-$ zero-shot)} & BGE & Expr.\ Acc & All Acc & Joint score & Quad.\ F1 & Joint F1 \\
\midrule
Qwen3.5-9B  &  \shortstack{+14.19\\{\scriptsize[+13.90, +14.48]}}  &  \shortstack{$-$0.11\\{\scriptsize[$-$1.04, +0.79]}}  &  \shortstack{$-$1.72\\{\scriptsize[$-$2.28, $-$1.10]}}  &  \shortstack{+8.11\\{\scriptsize[+6.83, +9.35]}}  &  \shortstack{+18.01\\{\scriptsize[+16.81, +19.23]}}  &  \shortstack{+4.45\\{\scriptsize[+3.62, +5.26]}} \\
Qwen3.8-27B  &  \shortstack{+15.72\\{\scriptsize[+15.43, +16.00]}}  &  \shortstack{+0.67\\{\scriptsize[$-$0.33, +1.66]}}  &  \shortstack{$-$1.48\\{\scriptsize[$-$2.06, $-$0.89]}}  &  \shortstack{+1.34\\{\scriptsize[+0.03, +2.72]}}  &  \shortstack{+3.14\\{\scriptsize[+2.10, +4.15]}}  &  \shortstack{+0.47\\{\scriptsize[$-$0.45, +1.38]}} \\
InternVL3.5-8B  &  \shortstack{+11.69\\{\scriptsize[+11.42, +11.95]}}  &  \shortstack{+9.16\\{\scriptsize[+8.21, +10.05]}}  &  \shortstack{+29.20\\{\scriptsize[+28.44, +29.98]}}  &  \shortstack{+30.85\\{\scriptsize[+29.67, +32.02]}}  &  \shortstack{+12.68\\{\scriptsize[+11.39, +13.83]}}  &  \shortstack{+12.63\\{\scriptsize[+11.95, +13.34]}} \\
Ovis2.5-9B  &  \shortstack{+11.15\\{\scriptsize[+10.88, +11.43]}}  &  \shortstack{+7.19\\{\scriptsize[+6.25, +8.13]}}  &  \shortstack{+8.90\\{\scriptsize[+8.22, +9.57]}}  &  \shortstack{+13.02\\{\scriptsize[+11.81, +14.22]}}  &  \shortstack{+15.49\\{\scriptsize[+14.45, +16.50]}}  &  \shortstack{+9.32\\{\scriptsize[+8.49, +10.23]}} \\
GLM-4.1V-9B  &  \shortstack{+10.80\\{\scriptsize[+10.48, +11.08]}}  &  \shortstack{+2.59\\{\scriptsize[+1.62, +3.59]}}  &  \shortstack{$-$0.45\\{\scriptsize[$-$1.10, +0.15]}}  &  \shortstack{+5.15\\{\scriptsize[+3.75, +6.41]}}  &  \shortstack{+49.02\\{\scriptsize[+47.58, +50.38]}}  &  \shortstack{+9.38\\{\scriptsize[+8.38, +10.36]}} \\
MiniCPM-V-4.5  &  \shortstack{+10.94\\{\scriptsize[+10.64, +11.23]}}  &  \shortstack{+4.94\\{\scriptsize[+4.16, +5.75]}}  &  \shortstack{+2.33\\{\scriptsize[+1.75, +2.91]}}  &  \shortstack{+22.81\\{\scriptsize[+21.71, +23.91]}}  &  \shortstack{+16.90\\{\scriptsize[+15.96, +17.86]}}  &  \shortstack{+9.68\\{\scriptsize[+9.01, +10.33]}} \\
\bottomrule
\end{tabular}
\caption{Effect of LoRA SFT on the 100 adaptation images, per model family: SFT minus zero-shot with paired image-bootstrap 95\% CIs.}
\label{tab:family_sft}
\end{table*}

\subsection{Proprietary Models on a Paired Subset}
\label{sec:api}

GPT-5.5 and Gemini-3.6-Flash are queried at temperature zero through a third-party OpenAI-compatible endpoint.
The fixed subset contains 100 test images and 160 opinions, separate from the adaptation images.
Table~\ref{tab:api100} compares them with open models on exactly the same queries.
A coordinate pilot indicated that Gemini returns boxes in $y$-first order, which is used for parsing.
The endpoint does not pin a model snapshot.
The two proprietary models recover all three Probe C phrases on at least 99.38\% of opinions.
Their joint scores are 20.00 and 28.75, showing a large gap after spatial and Emotion criteria are applied.
In Probe D, both also exceed 57 semantic F1 but remain below 10 joint F1.
The gap therefore recurs across supplied and discovered opinions on the paired subset.

\begin{table*}[tp]
\centering
\scriptsize
\setlength{\tabcolsep}{3.5pt}
\resizebox{\textwidth}{!}{%
\begin{tabular}{l cccc c ccc cccc}
\toprule
& \textbf{Probe A} & \multicolumn{4}{c}{\textbf{Probe B} (Acc@0.5)} & \multicolumn{3}{c}{\textbf{Probe C}} & \multicolumn{4}{c}{\textbf{Probe D}} \\
\cmidrule(lr){2-2} \cmidrule(lr){3-6} \cmidrule(lr){7-9} \cmidrule(lr){10-13}
\textbf{Model} & BGE & H & T & E & All & Joint score & Emo. & E box & \#Pred & Quad.\ F1 & Joint F1 & E gR \\
\midrule
\multicolumn{13}{l}{\textit{Proprietary (zero-shot)}} \\
GPT-5.5 & 65.70 & 74.38 & 29.03 & 8.75 & 37.47 & 20.00 & 76.25 & 11.25 & 271 & 57.08 & 6.03 & 3.75 \\
Gemini-3.6-Flash & 63.15 & 68.13 & 30.97 & 10.00 & 36.42 & \textbf{28.75} & \textbf{85.62} & 10.00 & 259 & 61.58 & 9.07 & 3.12 \\
\midrule
\multicolumn{13}{l}{\textit{Open, zero-shot}} \\
Qwen3.5-9B & 64.99 & \textbf{76.25} & 32.26 & \textbf{19.38} & \textbf{42.74} & 23.13 & 80.00 & 14.37 & 346 & 48.62 & 9.49 & 5.63 \\
Qwen3.8-27B & 63.79 & 75.00 & \textbf{32.90} & 17.50 & 41.89 & \textbf{28.75} & 80.63 & 14.37 & 188 & \textbf{62.07} & \textbf{10.92} & \textbf{7.50} \\
InternVL3.5-8B & 65.48 & 10.00 & 10.97 & 2.50 & 7.79 & 3.75 & 81.25 & 3.75 & 266 & 48.36 & 0.94 & 3.75 \\
Ovis2.5-9B & \textbf{67.09} & 61.25 & 20.65 & 7.50 & 29.89 & 18.12 & 83.75 & \textbf{15.00} & 156 & 51.90 & 7.59 & 4.37 \\
Ovis2.5-2B & 65.26 & 58.75 & 25.81 & 14.37 & 33.05 & 11.87 & 82.50 & 7.50 & 286 & 60.99 & 6.73 & 6.25 \\
\midrule
\multicolumn{13}{l}{\textit{Open, adapted on the 100 adaptation images}} \\
Qwen3.5-9B~+SFT & 81.42 & 75.00 & 34.19 & 20.62 & 43.37 & 36.88 & 89.38 & 26.25 & 214 & 68.45 & 15.51 & 18.12 \\
\bottomrule
\end{tabular}}
\caption{Paired comparison on 100 test images with 160 reference opinions.
Each model receives 475 queries for each of Probes A and B, 160 for C, and 100 for D.
The listed open models are rescored on exactly the same queries.
Both proprietary models are queried at temperature zero through an OpenAI-compatible endpoint.
Gemini boxes are read in the $y$-first order selected by a coordinate pilot.
E box is the Probe~C Expression box accuracy at IoU$\geq0.5$.
Bold marks the best zero-shot system.
}
\label{tab:api100}
\end{table*}

\section{Additional Adaptation Experiments}
\subsection{Seed Variance and Data Size}

\label{sec:appendix_robustness}
For Qwen3.5-9B, we repeat the 100-image adaptation with two further seeds and also train on nested 25- and 50-image subsets of the same images (Table~\ref{tab:adaptation}).
The main-text tables report seed 42.
The three 100-image runs differ by at most 1.9 points on Probe~C joint score and 1.7 points on joint opinion F1.
Probes B and C reach similar scores with 25 images, and Probe A is within 1.6 points of the 100-image result.
Probe D semantic F1 increases from 46.27 to 53.33 and 63.34 with 25, 50, and 100 images.
Parse rate increases from 78.28\% to 84.82\% and 95.89\%.
The data-size response is strongest for image-only discovery.
Expression accuracy varies by at most 1.16 points across the three sample sizes and by 1.12 points across the 100-image seeds.
The bottom block of Table~\ref{tab:adaptation} compares the seed-42 run with the zero-shot 9B and 27B models.

\begin{table*}[tp]
\centering
\footnotesize
\setlength{\tabcolsep}{3.4pt}
\resizebox{\textwidth}{!}{%
\begin{tabular}{l c cc c c ccccc}
\toprule
& & \textbf{Probe A} & \multicolumn{2}{c}{\textbf{Probe B}} & \textbf{Probe C} & \multicolumn{5}{c}{\textbf{Probe D}} \\
\cmidrule(lr){3-3} \cmidrule(lr){4-5} \cmidrule(lr){6-6} \cmidrule(lr){7-11}
\textbf{Qwen3.5-9B} & \textbf{Images} & BGE & E & All & Joint score & \#Pred & Quad.\ F1 & Joint F1 & E gR & Parse \\
\midrule
Zero-shot & 0 & 65.38 & 15.69 & 47.64 & 31.50 & 15,689 & 45.33 & 11.21 & 6.45 & 98.72 \\
+SFT & 25 & 78.01 & 15.06 & 45.90 & 40.85 & 13,886 & 46.27 & 11.24 & 10.43 & 78.28 \\
+SFT & 50 & 79.21 & 16.22 & 46.19 & 37.83 & 12,989 & 53.33 & 12.67 & 10.32 & 84.82 \\
+SFT, seed 42 & 100 & 79.57 & 15.58 & 45.92 & 39.61 & 11,991 & 63.34 & 15.66 & 14.73 & 95.89 \\
+SFT, seed 1 & 100 & 79.38 & 16.20 & 45.36 & 40.57 & 13,517 & 59.13 & 13.92 & 14.04 & 97.49 \\
+SFT, seed 2 & 100 & 79.46 & 15.08 & 45.41 & 38.67 & 12,664 & 61.05 & 14.49 & 15.34 & 95.68 \\
+SFT, mean $\pm$ s.d. & 100 & 79.47$\pm$0.10 & 15.62$\pm$0.56 & 45.56$\pm$0.31 & 39.62$\pm$0.95 & 12,724 & 61.17$\pm$2.11 & 14.69$\pm$0.89 & 14.70$\pm$0.65 & 96.35$\pm$0.99 \\
\bottomrule
\end{tabular}}
\medskip
\small
\begin{tabular}{lcc}
\toprule
Seed-42 SFT minus & Qwen3.5-9B zero-shot & Qwen3.8-27B zero-shot \\
\midrule
A: BGE cosine & +14.19 {\scriptsize[+13.90, +14.48]} & +15.30 {\scriptsize[+15.01, +15.59]} \\
B: Expression accuracy & $-$0.11 {\scriptsize[$-$1.04, +0.79]} & +0.60 {\scriptsize[$-$0.42, +1.61]} \\
B: Overall accuracy & $-$1.72 {\scriptsize[$-$2.28, $-$1.10]} & $-$2.10 {\scriptsize[$-$2.72, $-$1.47]} \\
C: Joint score & +8.11 {\scriptsize[+6.83, +9.35]} & +1.31 {\scriptsize[$-$0.01, +2.68]} \\
D: Semantic F1 & +18.01 {\scriptsize[+16.81, +19.23]} & +2.10 {\scriptsize[+1.08, +3.14]} \\
D: Joint F1 & +4.45 {\scriptsize[+3.62, +5.26]} & +0.16 {\scriptsize[$-$0.76, +1.07]} \\
D: Expression grounded recall & +8.28 {\scriptsize[+7.29, +9.25]} & +6.01 {\scriptsize[+4.98, +7.03]} \\
\bottomrule
\end{tabular}
\caption{Qwen3.5-9B adaptation with different sample sizes and seeds.
The upper panel reports individual runs and the mean and standard deviation across three 100-image seeds.
The lower panel compares 100-image Qwen3.5-9B SFT at seed 42 with two zero-shot baselines.
Brackets give paired 95\% image-bootstrap intervals over 2,000 resamples.
E gR denotes grounded Expression recall.}
\label{tab:adaptation}
\end{table*}

\subsection{Other Uses of the 100 Images}

\label{sec:alternatives}

\begin{table*}[!t]
\centering
\small
\setlength{\tabcolsep}{2.6pt}
\begin{tabular}{l cc c c ccc}
\toprule
& \textbf{A} & \multicolumn{2}{c}{\textbf{B}} & \textbf{C} & \multicolumn{3}{c}{\textbf{D}} \\
\cmidrule(lr){2-2} \cmidrule(lr){3-4} \cmidrule(lr){5-5} \cmidrule(lr){6-8}
 & BGE & Expr. & All & Joint & Quad. F1 & Joint F1 & Parse \\
\midrule
\multicolumn{8}{l}{\textbf{Qwen3.5-9B}} \\
Zero-shot & 65.38 & 15.69 & 47.64 & 31.50 & 45.33 & 11.21 & 98.72 \\
\midrule
\multicolumn{8}{l}{\textit{In-context demonstrations from the 100 images}} \\
$k=0$ (adapted prompt only) & 65.40 & 15.23 & 47.54 & 32.41 & 46.41 & 11.77 & 98.56 \\
$k=1$ & 70.82 & 15.48 & 46.30 & 25.62 & 49.51 & 9.99 & 97.68 \\
$k=3$ & 71.20 & 14.18 & 44.69 & 34.67 & 53.46 & 11.15 & 97.61 \\
\midrule
\multicolumn{8}{l}{\textit{LoRA SFT on the 100 images}} \\
All probes, LM only (default) & 79.57 & 15.58 & 45.92 & 39.61 & 63.34 & 15.66 & 95.89 \\
All probes, LM + vision tower & 79.28 & 17.15 & 46.53 & 41.03 & 60.86 & 15.11 & 93.03 \\
Probe~B data only & -- & 16.87 & 46.62 & -- & -- & -- & -- \\
Probe~D data only & -- & -- & -- & -- & 59.64 & 11.99 & 83.32 \\
\midrule
\multicolumn{8}{l}{\textit{Weight merging and rejection sampling}} \\
Seed soup (3 seeds) & 79.61 & 17.65 & 46.86 & 41.93 & 61.19 & 15.39 & 92.82 \\
Default + B-only (0.5/0.5) & 78.45 & 16.92 & 47.23 & 42.14 & 52.31 & 15.06 & 94.92 \\
TIES (default, B-only, D-only) & 79.39 & 14.31 & 44.09 & 38.70 & 59.94 & 14.54 & 94.30 \\
Rejection-sampling FT & 79.50 & 16.08 & 47.18 & 40.32 & 70.30 & 16.86 & 99.77 \\
Rejection-sampling FT, round 2 & 79.64 & 16.31 & 46.39 & 40.71 & 70.65 & 16.16 & 99.96 \\
\midrule
\multicolumn{8}{l}{\textbf{Qwen3.8-27B}} \\
Zero-shot & 64.27 & 14.98 & 48.03 & 38.29 & 61.24 & 15.50 & 100.00 \\
In-context, $k=1$ & 70.49 & 16.17 & 48.89 & 36.57 & 49.72 & 14.86 & 100.00 \\
In-context, $k=3$ & 72.06 & 16.03 & 48.77 & 41.54 & 56.89 & 16.20 & 99.98 \\
LoRA SFT, all probes (default) & 79.99 & 15.65 & 46.54 & 39.64 & 64.38 & 15.97 & 99.65 \\
\midrule
\multicolumn{8}{l}{\textbf{InternVL3.5-8B}} \\
Zero-shot & 65.51 & 3.24 & 9.84 & 3.94 & 47.55 & 1.05 & 99.49 \\
In-context, $k=1$ & 70.43 & 8.97 & 34.21 & 23.47 & 50.26 & 7.59 & 99.01 \\
In-context, $k=3$ & 69.88 & 7.05 & 28.36 & 32.94 & 59.01 & 11.98 & 99.59 \\
LoRA SFT, all probes (default) & 77.20 & 12.41 & 39.04 & 34.80 & 60.23 & 13.69 & 94.53 \\
\bottomrule
\end{tabular}
\caption{Alternative uses of the 100 adaptation images.
Columns report Probe A cosine, Probe B box accuracy, Probe C joint score, and Probe D semantic F1, joint F1, and parse rate.
All values are on a 0--100 scale.
Runs use seed 42 unless noted.}
\label{tab:alternatives}
\end{table*}

\paragraph{In-context demonstrations.}
We use the first $k$ adaptation images as demonstrations under the adapted prompt.
The same examples are fixed across queries.
Appendix~\ref{sec:adaptation_controls} examines the corresponding zero-demonstration prompt control.
Three demonstrations raise Probe A cosine to 71.20 and Probe D semantic F1 to 53.46.
Both remain below SFT, at 79.57 and 63.34.
Joint opinion F1 is 11.15, and Expression localization is 14.18\%.

For Qwen3.8-27B, three demonstrations raise Probe A cosine to 72.06, below SFT at 79.99.
Probe C joint score reaches 41.54 and Probe D joint F1 reaches 16.20.
These scores are comparable to the SFT results of 39.64 and 15.97.
Semantic discovery F1 instead falls from 61.24 zero-shot to 56.89 with three demonstrations.
For InternVL3.5-8B, one demonstration raises Holder localization from 9.50\% to 63.55\%.
Three demonstrations yield Probe C joint score of 32.94 and semantic discovery F1 of 59.01.
Expression localization remains below SFT, at 7.05\% versus 12.41\%.
Demonstrations can improve guided grounding while lowering semantic discovery, as the Qwen3.8-27B results show.
For InternVL, one example helps localization, while SFT yields the best Expression accuracy.

\paragraph{Training variants.}
Adding vision-encoder LoRA raises Probe B Expression accuracy to 17.15\%.
The 1.45-point gain over zero-shot has a paired interval excluding zero.
Semantic discovery F1 falls to 60.86 and parse rate to 93.03\%, both below default SFT.
Updating the visual encoder therefore trades a small localization gain for lower discovery scores in this run.
Probe B-only training reaches similar Expression accuracy, at 16.87\%.
Training only on localization also leaves Expression accuracy below 17\%.
Probe D-only training reaches joint F1 of 11.99, compared with 15.66 for all-probe training.
The D-only run also receives fewer updates than all-probe training.

\paragraph{Adapter merging.}
A uniform model soup averages LoRA updates from the three default seeds~\citep{wortsman2022soups}.
It reaches 41.93 on Probe C and 17.65\% on Probe B Expression, but lowers parse rate to 92.82\%.
Equal-weight merging of the default and Probe B-only adapters reaches 47.23\% overall Probe B accuracy.
It produces 16,009 opinions in Probe D, with semantic F1 of 52.31.
TIES merging~\citep{yadav2023ties} combines the default, Probe B-only, and Probe D-only adapters.
This variant scores below the default on every probe.
Merging shifts the balance between components.
The seed soup improves A, B, and C, while default SFT performs better on discovery.

\begin{table*}[!t]
\centering
\scriptsize
\setlength{\tabcolsep}{2.0pt}
\resizebox{\textwidth}{!}{%
\begin{tabular}{l cc cc cc cc cc}
\toprule
\textbf{Text from} $\rightarrow$ & \multicolumn{2}{c}{Qwen3.5-9B} & \multicolumn{2}{c}{Qwen3.8-27B} & \multicolumn{2}{c}{Qwen3.5-9B~+SFT} & \multicolumn{2}{c}{Qwen3.5-9B~+RFT} & \multicolumn{2}{c}{InternVL~+SFT} \\
\cmidrule(lr){2-3} \cmidrule(lr){4-5} \cmidrule(lr){6-7} \cmidrule(lr){8-9} \cmidrule(lr){10-11}
\textbf{Boxes from} & F1 & E gR & F1 & E gR & F1 & E gR & F1 & E gR & F1 & E gR \\
\midrule
Original VLM boxes         & 11.21 & 6.45 & 15.50 & 8.71 & 15.66 & 14.73 & 16.86 & 14.97 & 13.69 & 12.48 \\
G-DINO, top-1           & 11.58 & 9.71 & 14.94 & 8.56 & 15.17 & 12.73 & 15.19 & 12.94 & 13.67 & 12.06 \\
G-DINO, snap            & 13.11 & 9.48 & 17.02 & 9.54 & 18.42 & 16.95 & 18.81 & 16.97 & 16.22 & 15.22 \\
G-DINO, snap, IoU$\geq$0.5 & 12.35 & 6.48 & 16.67 & 8.87 & 16.66 & 14.84 & 17.50 & 15.17 & 14.72 & 12.91 \\
G-DINO~+FT, top-1       & 11.75 & 10.22 & 15.10 & 8.99 & 15.74 & 13.48 & 15.82 & 13.78 & 14.15 & 12.42 \\
G-DINO~+FT, snap        & 13.07 & 9.31 & 16.91 & 9.36 & 18.31 & 16.78 & 18.76 & 16.64 & 15.99 & 14.52 \\
\bottomrule
\end{tabular}}
\caption{Extract-then-ground for Probe~D: joint opinion F1 and grounded Expression recall (E gR) when Grounding DINO re-boxes the VLM phrases.}
\label{tab:pipeline}
\end{table*}

\begin{table*}[!t]
\centering
\small
\setlength{\tabcolsep}{2.6pt}
\begin{tabular*}{\textwidth}{@{\extracolsep{\fill}}lrrrrrrr@{}}
\toprule
& \multicolumn{2}{c}{\textbf{A} (Expr.\ BGE)} & \multicolumn{2}{c}{\textbf{D} found} & \multicolumn{2}{c}{\textbf{D} joint} & \textbf{D} all \\
\cmidrule(lr){2-3} \cmidrule(lr){4-5} \cmidrule(lr){6-7}
\textbf{Model} & Expl. & Impl. & Single & Multi & Single & Multi & persp. \\
\midrule
Qwen3.5-9B                   & 65.67 & 60.18 & 74.5 & 66.2 & 20.4 & 10.6 & 55.1 \\
Qwen3.8-27B                  & 67.33 & 63.63 & 73.1 & 56.2 & 19.8 & 10.5 & 37.5 \\
Qwen3.5-9B~+SFT              & 82.43 & 72.52 & 89.3 & 71.4 & 23.9 & 12.2 & 56.8 \\
Qwen3.8-27B~+SFT             & 81.92 & 72.89 & 91.3 & 74.8 & 24.7 & 12.5 & 60.7 \\
Ovis2.5-9B~+SFT              & 77.38 & 72.15 & 93.5 & 68.7 & 23.6 & 10.9 & 49.9 \\
GLM-4.1V-9B-Thinking~+SFT    & 81.12 & 72.31 & 92.5 & 71.4 & 23.5 & 13.1 & 54.6 \\
\bottomrule
\end{tabular*}
\caption{Full property analysis corresponding to Figure~\ref{fig:property_summary}.
Probe A reports Expression cosine ($\times100$) for explicit and implicit cues in 599 sampled opinions.
Probe D reports semantic recall (found) and joint recall on all test images, split by single-holder and multi-holder scenes.
The final column gives the percentage of multi-holder scenes with every perspective found.
Property definitions and sampling appear in Appendix~\ref{sec:appendix_statistics}.}
\label{tab:properties}
\end{table*}

\paragraph{Rejection-sampling fine-tuning (RFT).}
\label{sec:rft}
RFT augments the 1,868 training queries from the same 100 adaptation images.
The default SFT model, Qwen3.5-9B, answers each query once greedily and eight times at temperature 1.0~\citep{yuan2023scaling}.
We keep up to two accepted answers per query, excluding the reference answer.

Acceptance uses the training annotations and the phrase and box metrics in Appendix~\ref{sec:appendix_metrics}.
Probe A requires phrase cosine $\geq0.85$, and Probe B requires box IoU $\geq0.7$.
Probe C requires all three phrase similarities to reach 0.5, exact Emotion, and mean box IoU $\geq0.5$ over roles with reference boxes.
Probe D requires joint opinion F1 $\geq0.5$, computed for the current training image using the joint matching rule in Appendix~\ref{sec:appendix_metrics}.
Its predicted opinion count must also be no greater than the reference count.

The training set combines 1,868 reference targets and 1,619 accepted answers.
We train a new LoRA adapter from the base model using the configuration in Appendix~\ref{sec:appendix_reproducibility}.
This includes three epochs, learning rate $10^{-4}$, and effective batch size 16.
The expanded answer set increases training exposure within the same 100-image partition.
We evaluate the final checkpoint with greedy decoding.

Semantic discovery F1 reaches 70.30, joint F1 reaches 16.86, and parse rate reaches 99.77\%.
The model produces 10,373 opinions, compared with 11,991 after default SFT.
This reduction is consistent with the opinion-count constraint used to select training answers.
Expression localization remains at 16.08\%, compared with 15.58\% after SFT.
The main gain is semantic discovery, with little change in cue localization.

\paragraph{Second RFT round.}
A second configuration samples answers from the first-round RFT model and lowers the Probe B acceptance threshold from IoU 0.7 to 0.5.
We again train LoRA from the base model on the reference targets plus 1,969 accepted answers.
Semantic F1 rises from 70.30 to 70.65, while joint F1 falls from 16.86 to 16.16.
Expression localization changes from 16.08\% to 16.31\%.
This configuration improves semantic agreement slightly, with no gain in joint opinion recovery.

\definecolor{roleH}{HTML}{4A3AA7}
\definecolor{roleT}{HTML}{1B9E77}
\definecolor{roleE}{HTML}{E8590C}
\newcommand{\casecell}[3]{\begin{minipage}[t]{0.485\textwidth}\centering#1\par\vspace{3pt}\raggedright\footnotesize #2\par\vspace{3pt}{\footnotesize\setlength{\tabcolsep}{1.6pt}\begin{tabular}{@{}l>{\raggedright\arraybackslash}p{0.2\linewidth}>{\raggedright\arraybackslash}p{0.22\linewidth}>{\raggedright\arraybackslash}p{0.3\linewidth}l@{}}
\toprule
 & \textcolor{roleH}{\textbf{Holder}} & \textcolor{roleT}{\textbf{Target}} & \textcolor{roleE}{\textbf{Expression}} & \textbf{Emo.} \\
\midrule
#3
\bottomrule
\end{tabular}}\end{minipage}}

\begin{figure*}[!t]
\centering
\textbf{Probe A: region interpretation}\par\smallskip
\casecell{\includegraphics[height=3.5cm,width=\linewidth,keepaspectratio]{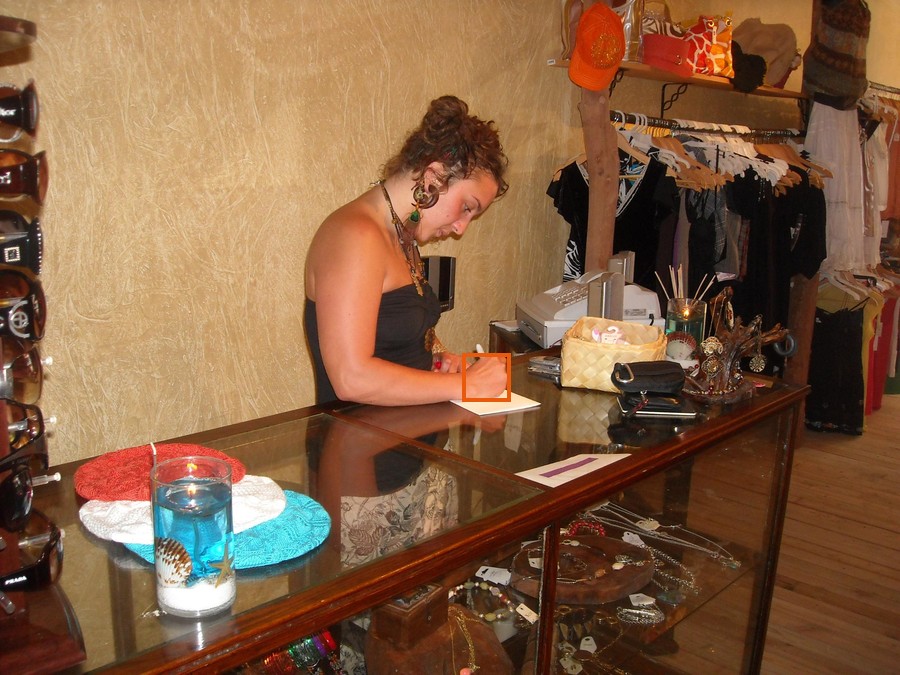}}{\textbf{Higher cosine.} Input: the Expression box. Cosine 0.85.}{%
Ref & shopkeeper & notes & looks focused while writing down & Pos \\
Pred & -- & -- & focused gaze toward the paper & -- \\}\hfill
\casecell{\includegraphics[height=3.5cm,width=\linewidth,keepaspectratio]{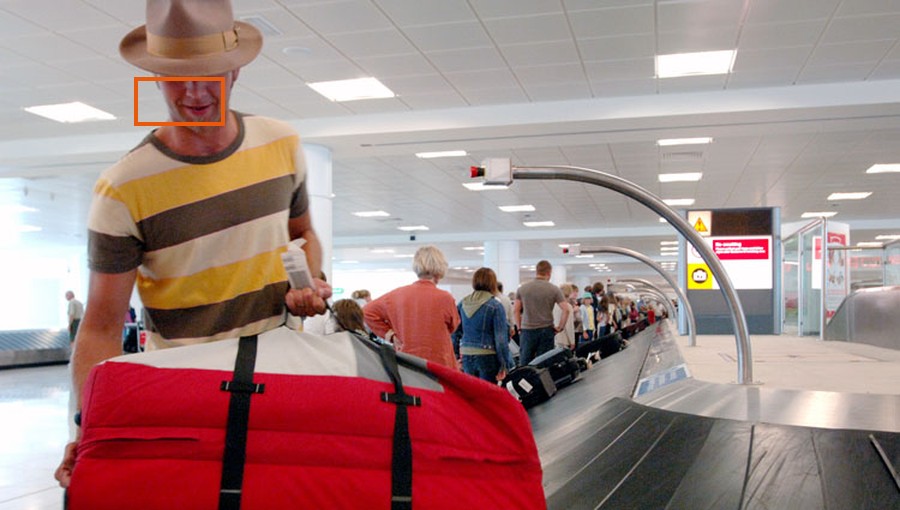}}{\textbf{Lower cosine.} Input: the Expression box. Cosine 0.45.}{%
Ref & man & luggage & smiles warmly at & Pos \\
Pred & -- & -- & man pulling large red suitcase & -- \\}
\par\bigskip
\textbf{Probe B: evidence localization}\par\smallskip
\casecell{\includegraphics[height=3.5cm,width=\linewidth,keepaspectratio]{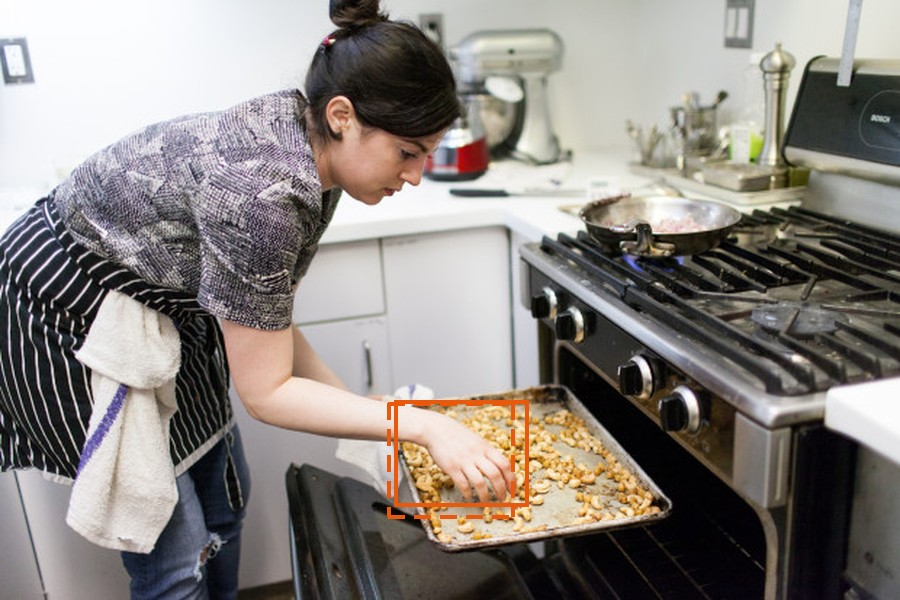}}{\textbf{Success.} Input: the Expression phrase. The predicted box covers the hands and the tray (IoU 0.76).}{%
Ref & woman & tray of nuts & carefully taking out & Pos \\
Pred & -- & -- & box only & -- \\}\hfill
\casecell{\includegraphics[height=3.5cm,width=\linewidth,keepaspectratio]{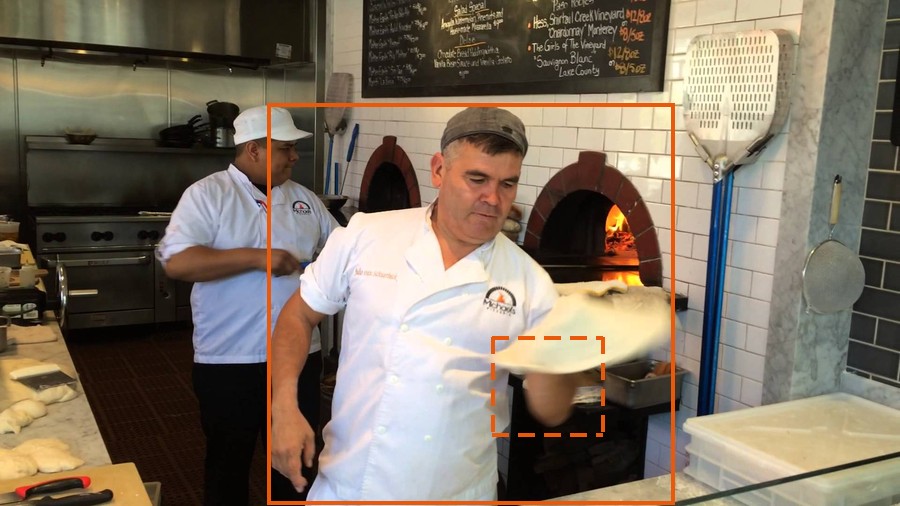}}{\textbf{Failure.} Input: the Expression phrase. The prediction covers the whole chef (IoU 0.07).}{%
Ref & chef & pizza dough & skillfully tossing & Pos \\
Pred & -- & -- & box only & -- \\}
\caption{Component-level examples from zero-shot Qwen3.5-9B.
Probe A compares higher and lower phrase cosine.
Probe B compares a correct and an incorrect box at IoU $\geq0.5$.
Orange boxes mark Expression regions.
Solid boxes are inputs in A and predictions in B, where dashed boxes mark references.
Ref lists the full reference opinion for context.
Dashes in Pred mark fields that the probe does not predict.}
\label{fig:case_study}
\end{figure*}

\subsection{Prompt and Output Conditions}
\label{sec:adaptation_controls}

\paragraph{Holding the prompt fixed.}
The Qwen3.5-9B prompt control uses the base model with the adapted prompt and no demonstrations (Table~\ref{tab:alternatives}).
Both the control and SFT use greedy decoding without thinking.
Changing only the prompt moves each probe score by at most 1.08 points.
Under the same prompt, SFT raises semantic F1 from 46.41 to 63.34 and joint F1 from 11.77 to 15.66, gains of 16.93 and 3.89 points.
Expression accuracy changes from 15.23\% to 15.58\%.
Discovery improves more than cue localization.

\paragraph{Discovery and output validity.}
Parse rate falls from 98.56\% to 95.89\% while both discovery scores improve.
The gain thus occurs despite fewer parseable outputs.
GLM instead improves discovery alongside a parse-rate increase from 74.37\% to 99.51\%.
Its inference also changes from sampled thinking to greedy direct answers (Appendix~\ref{sec:appendix_reproducibility}).

\begin{figure*}[!t]
\centering
\textbf{Probe C: guided opinion grounding}\par\smallskip
\casecell{\includegraphics[height=3.5cm,width=\linewidth,keepaspectratio]{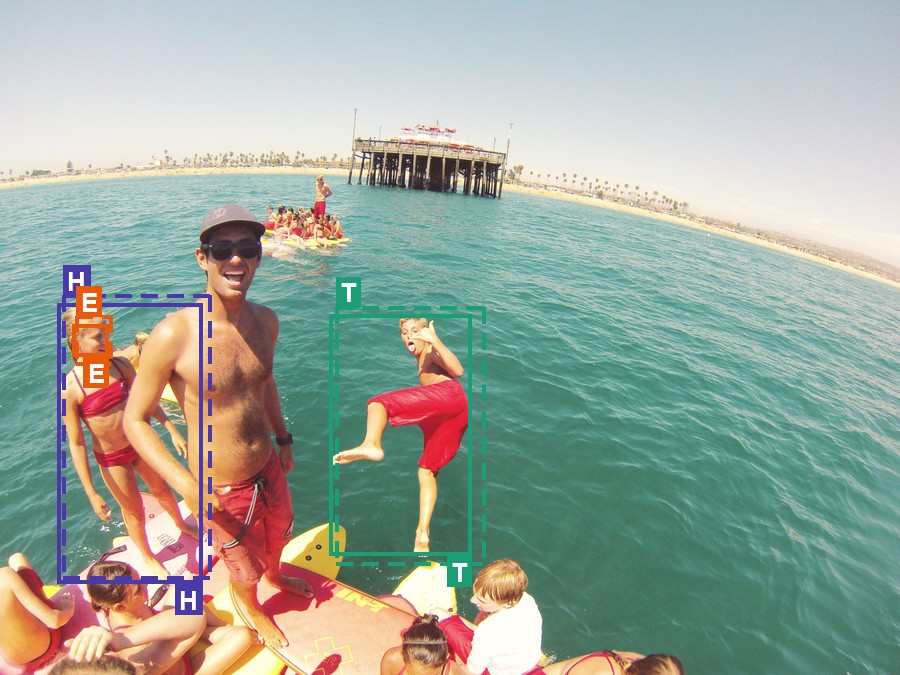}}{\textbf{Success.} ``The girl in pink is happily looking at the boy jumping into the water.'' Cosine 1.00 for H, 0.88 for T, and 0.77 for E. Mean IoU 0.77.}{%
Ref & girl in pink & boy jumping into the water & happily looking at & Pos \\
Pred & Girl in pink & Boy jumping & Smiling face & Pos \\}\hfill
\casecell{\includegraphics[height=3.5cm,width=\linewidth,keepaspectratio]{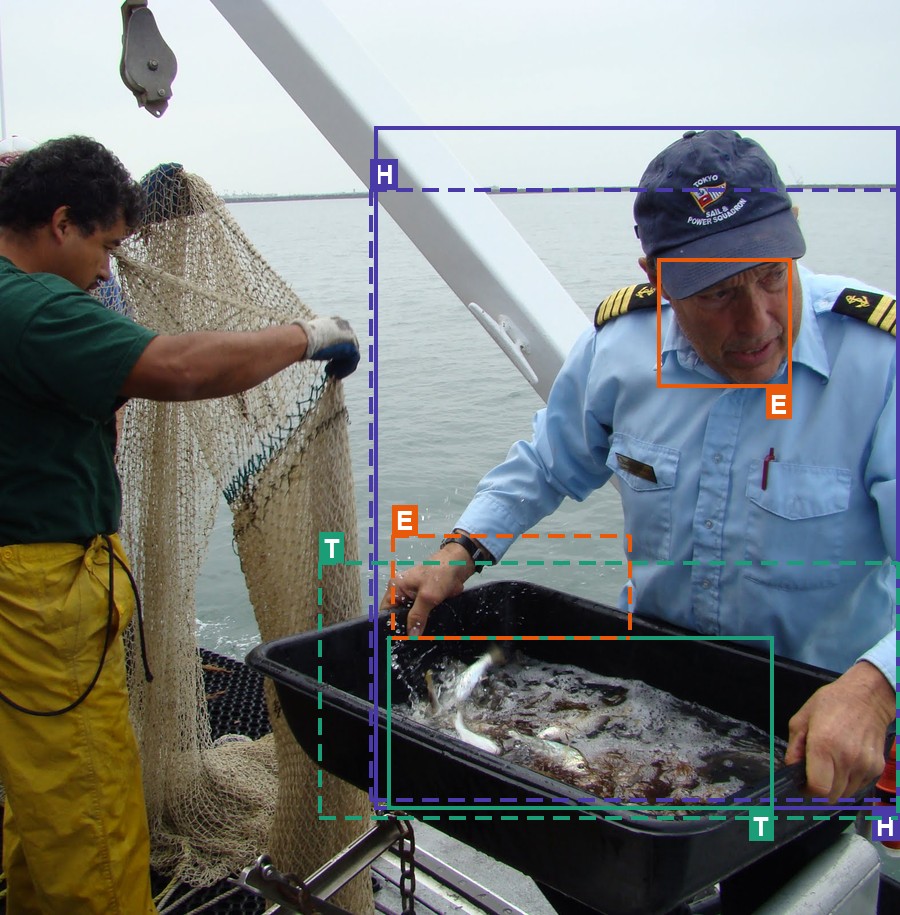}}{\textbf{Failure.} ``The man on the right carefully examines the fish in the container.'' All phrases pass. IoU is 0.88 for H, 0.45 for T, and 0.00 for E, so the mean of 0.45 fails the spatial gate.}{%
Ref & man on the right & fish in the container & carefully examines & Neu \\
Pred & man on the right & fish in the container & man looking down at the fish & Neu \\}
\par\bigskip
\textbf{Probe D: image-only opinion discovery}\par\smallskip
\casecell{\includegraphics[height=3.5cm,width=\linewidth,keepaspectratio]{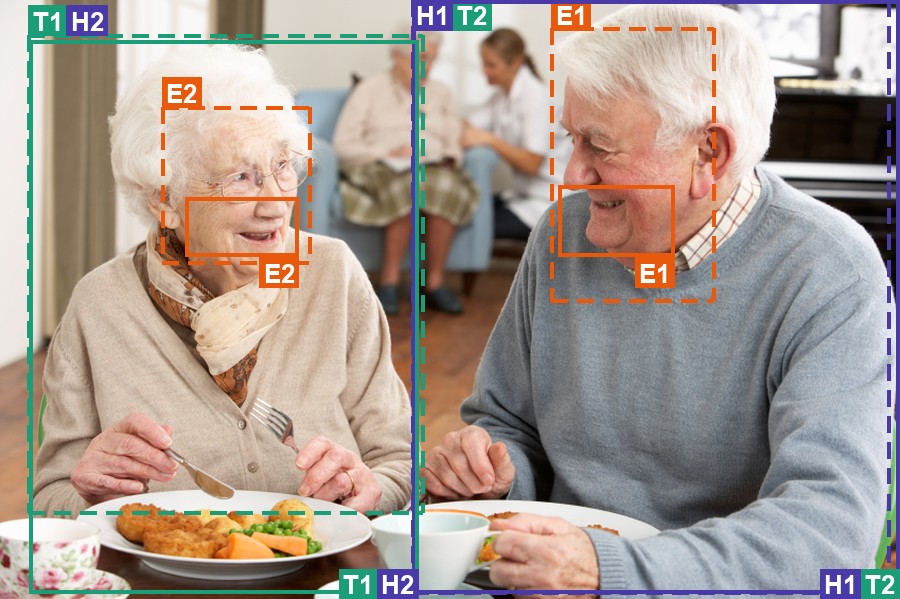}}{\textbf{Success.} Both opinions are joint matches (mean IoU 0.67 and 0.71).}{%
Ref 1 & elderly man & elderly woman & attentively engaging & Pos \\
Pred 1 & elderly man & elderly woman & smiling & Pos \\
\midrule
Ref 2 & elderly woman & elderly man & smiling warmly & Pos \\
Pred 2 & elderly woman & elderly man & smiling with open mouth & Pos \\}\hfill
\casecell{\includegraphics[height=3.5cm,width=\linewidth,keepaspectratio]{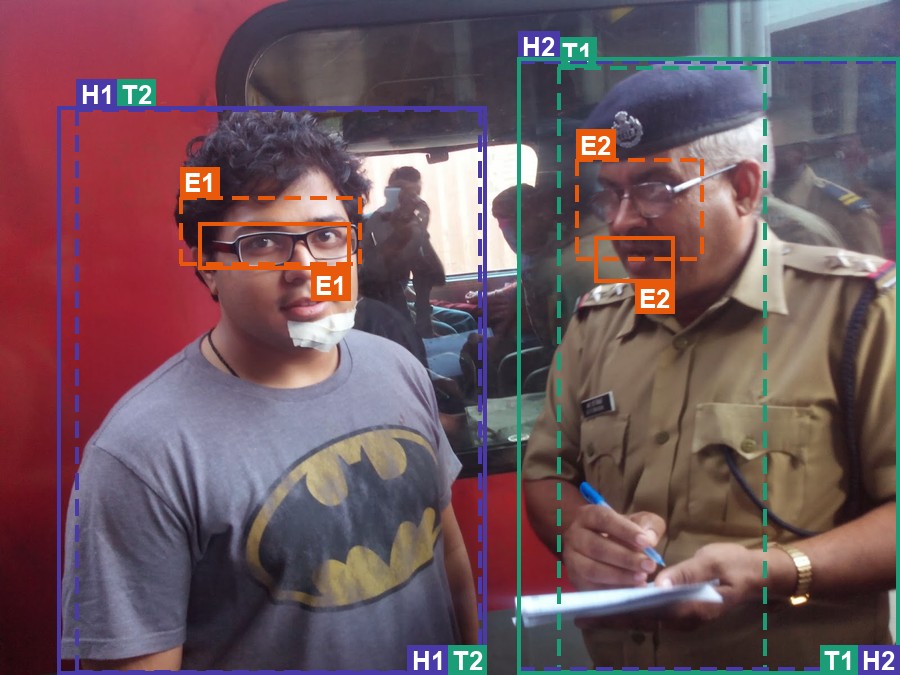}}{\textbf{Failure.} The boxes fit (mean IoU 0.65 and 0.68), but both emotions are wrong. No joint match.}{%
Ref 1 & the man & the police officer & looks worried about & Neg \\
Pred 1 & Young man & Police officer & Gaze directed at the police officer & Neu \\
\midrule
Ref 2 & the police officer & the man & shows concern for & Pos \\
Pred 2 & Police officer & Young man & Mouth open while speaking & Neu \\}
\caption{Joint matches (left) and failures (right) in Probes C and D for zero-shot Qwen3.5-9B.
Purple, green, and orange mark Holder, Target, and Expression.
Dashed boxes and Ref rows show references, while solid boxes and Pred rows show predictions.
Numbers in D identify opinions.
Joint matching requires phrase agreement, exact Emotion, and mean IoU $\geq0.5$ across available role boxes.}
\label{fig:case_study_cd}
\end{figure*}

\section{Additional Diagnostic Analyses}
\subsection{Detector-based Box Selection}
\label{sec:appendix_pipeline}

We hold the predicted Probe D phrases fixed and replace their boxes with Grounding DINO proposals.
Table~\ref{tab:pipeline} compares five text sources.
Selecting the top-scoring detection changes joint opinion F1 by $-1.67$ to $+0.37$ points.
The \emph{snap} variant selects the box with highest IoU to the original VLM box among the top 20 detections.
It improves joint F1 by 1.52--2.76 points across all five sources, with paired intervals excluding zero.

Snap reaches joint F1 of 18.81 for the rejection-sampling model and 18.42 for default adapted Qwen3.5-9B.
The 0.39-point difference has an interval including zero.
Grounded Expression recall rises from 14.97\% to 16.97\% and from 14.73\% to 16.95\%, respectively.
Snap places no lower bound on overlap with the original box.
It can therefore select a detection that overlaps that box only partly.

A thresholded variant accepts a detector box only when this overlap reaches IoU 0.5.
It yields joint F1 of 17.50 and 16.66 for the same two text sources.
Their gains of 0.64 and 0.99 points have paired intervals excluding zero.
Grounded Expression recall changes by only 0.21 and 0.11 points, with intervals including zero.
The original VLM box is useful as a proposal-selection cue, but a strict overlap threshold removes some beneficial corrections.
Fine-tuning the detector changes either variant by at most 0.7 points.

\subsection{Implicit Cues and Multiple Holders}
\label{sec:appendix_properties}

Table~\ref{tab:properties} provides exploratory results by opinion property.
Implicitness uses the sampled model-assisted labels in Appendix~\ref{sec:appendix_statistics}.
Multi-holder scenes contain at least two distinct annotated holders.
For the four adapted models shown, semantic opinion recovery is 89.3--93.5\% in single-holder scenes and 68.7--74.8\% in multi-holder scenes.
Joint opinion recovery is also lower in multi-holder scenes.
Among these adapted models, Ovis ranks first in single-holder semantic recall and last in multi-holder recall.
In Probe A, Expression cosine is lower for implicit than explicit cues for every model shown.
Both Qwen models improve more on explicit cues, widening the cosine gaps from 5.49 to 9.91 and from 3.70 to 9.03 points.
These comparisons use the model-assisted labels and naturally occurring scene groups described in Appendix~\ref{sec:appendix_statistics}.

\subsection{Region Size and Annotation Granularity}
\label{sec:region_extent}

\paragraph{Spatial extent of the cues.}
Median box area is 2.0\% of the image for Expressions and 17.7\% for Holders.
Expression and Holder boxes coincide for 1,605 of 7,150 test opinions, or 22.4\%.
Expression annotations thus include both compact cues and evidence covering a person.
Box extent follows the cue-specific rules in Appendix~\ref{sec:appendix_guidelines}.

\paragraph{Sensitivity to boundary displacement.}
Two square boxes of side length $l$, displaced horizontally by $d$ with $0\leq d<l$, have
\begin{equation}
\operatorname{IoU}=\frac{l-d}{l+d}.
\end{equation}
An IoU of at least 0.5 permits displacement up to $l/3$.
The same absolute offset therefore causes greater IoU loss for smaller boxes.
This example links box granularity with IoU sensitivity.

\section{Qualitative Examples}
\label{sec:appendix_cases}

Figures~\ref{fig:case_study} and~\ref{fig:case_study_cd} pair reference opinions with zero-shot Qwen3.5-9B outputs to illustrate phrase, box, and Emotion errors.
Probe A contrasts higher and lower cosine scores, while Probes B--D show passing and failing cases.

\paragraph{Component errors.}
In the lower-cosine Probe A example, the input box marks the face, but the answer describes the action of the whole person.
In the Probe B failure, the model boxes the whole chef for a hand action, although the zero-shot prompt asks for the cue region.

\paragraph{Joint evaluation.}
In the Probe C failure, all phrases pass the semantic threshold and Emotion matches.
The Expression box lies on the face while the reference marks the hands.
The mean IoU then falls to 0.45, below the spatial gate.
In the Probe D success, both opinions satisfy the joint criterion.
Their mean IoUs pass even though the Expression boxes cover only the mouths and have IoUs of 0.19 and 0.30.
The mean-IoU gate accepts these opinions despite their low Expression overlap, which the role-level scores capture.
In the Probe D failure, the model again finds both perspectives with fitting boxes, but labels both attitudes Neutral.
Both predictions fail on Emotion despite fitting boxes.

\FloatBarrier

\end{document}